\documentclass[journal]{IEEEtran}

\usepackage{amssymb}
\usepackage{bm}
\usepackage[ruled,norelsize]{algorithm2e}
\usepackage{amsmath,graphicx}
\usepackage{tabularx}
\usepackage[usenames, dvipsnames]{color}

\makeatletter
\newcommand{\removelatexerror}{\let\@latex@error\@gobble}
\makeatother

\begin{document}

\title{Tensor-Based Classification Models for Hyperspectral Data Analysis}
%
%
%

\author{Konstantinos Makantasis, Anastasios Doulamis, Nikolaos Doulamis and Antonis Nikitakis 
\thanks{Konstantinos Makantasis is with the KIOS Research and Innovation Center of Excellence, Cyprus.}
\thanks{Anastasios Doulamis and Nikolaos Doulamis are with the National Technical University of Athens, Greece.}
\thanks{Antonis Nikitakis is with Althexis Solutions Ldt, Cyprus.}
\thanks{This work was support by the European Union funded H2020 STOP-IT project, “Strategic, Tactical, Operational Protection of water Infrastructure against cyber-physical Threats,” grand agreement, 740610.}
}

%
%

\markboth{Journal of ,~Vol. , No. , ~2017}%
{}

\maketitle

\begin{abstract}
In this paper, we present tensor-based linear and nonlinear models for hyperspectral data classification and analysis. By exploiting principles of tensor algebra, we introduce new classification architectures, the weight parameters of which satisfy the {\it rank}-1 canonical decomposition property. Then, we propose learning algorithms to train both linear and non-linear classifiers. The advantages of the proposed classification approach are that i) it significantly reduces the number of weight parameters required to train the model (and thus the respective number of training samples), ii) it provides a physical interpretation of model coefficients on the classification output and iii) it retains the spatial and spectral coherency of the input samples. The linear tensor-based model exploits principles of logistic regression assuming the {\it rank}-1 canonical decomposition property among its weights. For the non-linear classifier, we propose a modification of a feedforward neural network (FNN), called {\it rank}-1 FNN, since its weights satisfy again the {\it rank}-1 canonical decomposition property. An appropriate learning algorithm is also proposed to train the network. Experimental results and comparisons with state of the art classification methods, either linear (e.g., Linear SVM) or non-linear (e.g., deep learning) indicates the outperformance of the proposed scheme, especially in cases where a small number of training samples is available.

\end{abstract}


\IEEEpeerreviewmaketitle

\section{Introduction}
\label{sec:intro}
\IEEEPARstart{T}{he} recent advances in optics and photonics have stimulated the deployment of hyperspectral imaging sensors of high spatial and spectral resolution. These sensors are now placed on satellite, unmanned aerial vehicle, and ground acquisition platforms used for material, object and terrain land detection and classification \cite{wycoff2013non}. Although high spatial and spectral resolution improves classification accuracy, it also imposes several research challenges derived as a consequence of the so-called "curse of dimensionality"; the difficulties arising when we need to analyze and organize data in high dimensional spaces. Hyperspectral data have their own unique characteristics, though being applied for a wide variety of applications, such as agriculture, surveillance, astronomy and biomedical imaging \cite{chang2013hyperspectral}; i) high dimensional data, ii) limited number of labeled samples and iii) large spatial variability of spectral signatures \cite{camps2005kernel}.      

Most of existing works, concerning hyperspectral image classification, follow the conventional workflow of pattern recognition process, consisting of two separate steps. First, features are extracted from the raw data, creating labelled training datasets. Second, classifiers, linear or non-linear, such as Support Vector Machines (SVM) and Neural Networks (NNs) \cite{camps2009kernel}, are used to map the extracted features to the target (desired) outputs. The key problem, however, in applying such conventional processes in classifying high dimensional hyperspectral data is that a large number of labelled training samples is required to model the statistical input diversities and consequently to well train the classifier. In remote sensing applications, collection of a large number of labelled data is an expensive and time consuming process. Another drawback is that classifiers are often used as "black boxes" \cite{benitez1997artificial}. This means that there is no a direct interpretation of how spatial and spectral bands contribute to the final classification outcome.       

One way to address issues deriving from the high dimensionality and heterogeneity of the data is to employ statistical learning methods \cite{camps2014advances}. However, even in this case, the problem of extracting a set of appropriate features remains. Features significantly affect the classification outcome and for data laying in high dimensional spaces is really an arduous task to estimate a suitable set of discriminant features so as to increase the accuracy of the classifier. 

For this reason, recently deep learning paradigms have been investigated for classifying hyperspectral data \cite{lecun1998gradient,hinton2006reducing,hinton2006fast,bengio2007greedy}. Deep learning machines receive as inputs, instead of features, the raw sensory data. Then, they non-linearly transform the raw inputs to hierarchies of representations which are used, in the following, as "the most suitable features" in a supervised mapping phase. 
Thus, deep learning tackles feature-related issues. This is also proven by the current research outcomes \cite{chen2014deep,Makantasis-etal:15,vakalopoulou2015building,papadomanolaki2016benchmarking} indicating the outperformance of deep learning machines in accurately detecting various objects in hyperspectral imaging data. Examples include the detection of man-made constructions rather than natural ones \cite{mnih2012learning,makantasis2015deepmm}, vehicles' detection \cite{orr2003neural}, object tracking \cite{Kandylakis}, land cover mapping \cite{kussul2016deep} and critical infrastructure assessment \cite{makantasis2015tunnel}. 

However, a typical deep learning architecture contains a huge number of tunable parameters implying that a large number of samples is also needed to accurately train the network. In addition, deep learning processes present high computational complexity.   

Tensor-based machine learning models are promising alternatives for hyperspectral data classification \cite{zhou2013tensor, tan2012logistic}. In particular, Zhou et al. \cite{zhou2013tensor} and Tan et. al \cite {tan2012logistic} have introduced a linear model using tensor-based regression with applications in neuro-imaging data analysis \cite{zhou2013tensor} and for classification \cite{tan2012logistic}. These aproaches are considered as the first works discussed the statistical inference procedure for the general raw tensor regression. In conventional learning models, usually the inputs are vector data. Therefore, in case of multidimensional input arrays, first tensor vectorization is carried out. However, vectorization destroys the inherent spatial and spectral structure of the input which can offer a physical interpretation of how spatial information and spectral bands contribute to the classification outcome. Furthermore, tensor vectorization fails to address issues stem from the high dimensionality of the data since again a large number of tunable parameters is required. To handle these limitations, we need to consider the input data as tensors, keeping the spatial and spectral structure of the data, and then, using principles of tensor algebra, to find out ways to reduce the number of parameters needed to be estimated during training.     

\subsection{Our Contribution}
In this work we propose a new machine learning model which receives as inputs multidimensional tensor signals as they are the hyperspectral images, i.e., 3D tensor cubes. The model weight parameters satisfy the {\it rank}-1 canonical decomposition property meaning that the weight parameters are decomposed as a linear combination of a minimal number of possibly non-orthogonal rank-1 terms \cite{de2004computation}. Using such decomposition of the model weights we are able to significantly reduce the number of parameters required to train the classifier. Thus, a smaller labelled dataset is needed than in conventional learning approaches where the tensor inputs are first vectorized.   

Another advantage of the {\it rank}-1 canonical decomposition property for the model weights is that it retains the structure of the spatial and spectral band information, which is a very important aspect for hyperspectral data classification. This is due to the fact that it actually permits the extraction of valuable information regarding the contribution of each of the hyperspectral bands to the classification. Thus, the proposed canonical decomposition provides a physical interpretation of the classification outcome, i.e., how the location of the pixels (spatial information) and the spectral bands (spectral information) influence the final classification performance.  

Our work is motivated from \cite{zhou2013tensor} which proposes a single linear output tensor regression model for binary classification. In contrast to \cite{zhou2013tensor}, in this paper, a multi-class classification problem is investigated using tensor-based logistic regression models of multiple outputs. In addition, the {\it rank}-1 canonical decomposition property is also applied, apart from high-order linear, to non-linear classifiers, which is not a straightforward process.  The proposed high order non-linear model is relied on a modification of a feedforward neural network (FNN), while it retains the universal approximation principles; capability of the network to approximate any unknown function, under some assumptions of continuity, within any degree of accuracy. The main difference is that the model weights satisfy the {\it rank}-1 canonical decomposition property. Therefore, the number of parameters (and consequently the number of training samples) are significantly reduced, especially for cases where tensor inputs are considered. A new learning algorithm is also introduced to train the network without spoiling the canonical decomposition assumption. We call the proposed high-order nonlinear model as {\it rank}-1-FNN. 

To sum up, the main contribution of this paper is threefold. First, we introduce new learning models, linear and non-linear, that consider a {\it rank}-1 canonical decomposition of their weight parameters. This way, a quite smaller number of weights is required compared to conventional learning paradigms. This, in the sequel, requires a smaller number of samples used to train the model, which is in the line of remote sensing applications, where a limited number of samples is available. The new models are suitable for tensor input data of high dimensions. Second, the {\it rank}-1 canonical decomposition property allows for a physical interpretation of how each spatial and spectral band of the tensor inputs affects the classification outcome. This is an important attribute in analyzing hyperspectral data. Third, the introduction of a {\it rank}-1 FNN non-linear classifier allows for modelling of complex relationships, due to the universal approximation property of neural nets \cite{csaji2001approximation}, while simultaneously keeping the aforementioned advantages.

The rest of this paper is organized as follows; Section \ref{sec:notation} presents the problem formulation, as well as, the notation and tensor algebra operations that will be used throughout the paper. Sections \ref{sec:high_order_linear} and \ref{sec:nonlinear} present the development of the high-order linear and nonlinear models. Experimental evaluation of the developed models is presented in Section \ref{sec:experiments} and Section \ref{sec:conclusions} concludes this work. 

\section{Problem Formulation and Tensor Algebra Notation}
\label{sec:notation}
\subsection{Problem Formulation}
\label{sec:problem_formulation}
Let us denote as $\bm X_i$ the $i$-th patch of a hyper-spectral image that we would like to classify into one of $C$ available classes. Each class expresses, for example, the type of vegetation or soil properties for the patch $\bm X_i$. In this paper, we consider $\bm X_i$ as a 3D tensor, i.e., $\bm X_i \in \mathbb R^{p_1 \times p_2\times p_3}$, where variables $p_1$ and $p_2$ refer to the spatial dimensions of the hyperspectral patch and $p_3$ to the number of spectral bands. In a more general case, variable $\bm X_i$ can be seen as a $D$-dimensional tensor, that is,  $\bm X_i \in \mathbb R^{p_1 \times \cdots \times p_D}$.

Let us also denote as $p^k_w(\bm X_i)$, with $k=1,\cdots,C$ a relationship (linear or non-linear) that expresses the probability of the observation $\bm X_i$ to belong to the $k$-th class. Subscript $w$ indicates dependence of the relationship $p^k_w(\cdot)$ on weight parameters. Aggregating the values $p^k_w(\cdot)$ over all $C$ classes, we form a classification vector, say $\bm y_i$, the elements of which $y_{i,k} \equiv p^k_w(\cdot)$. Then, the maximum probability value over all $k$ classes indicates the class to which the hyperspectral image patch belongs to.        
\begin{equation}
	\hat k=\arg\max_{\forall k} \{{ y_{i,k} \equiv p^k_w(\bm X_i)}\}.
\end{equation}

The values of $y_{i,k}$ are estimated using machine learning algorithms. For this reason, a training dataset consisting $N$ samples is considered
\begin{equation}
\label{eq:dataset}
	\mathcal S = \{(\bm X_i, \bm t_i)\}_{i=1}^N ,
\end{equation}
where $\bm t_i$ is a $C$-dimensional vector, the elements of which $t_{i,j}$ are all zero except for one which equals unity indicating the class to which  $\bm X_i$ belongs to. That is, $\bm t_i \in \{0,1\}^C$ and $\sum_{j=1}^C t_{i,j} = 1$. In the following we omit subscript $i$ for simplicity purposes if we refer to an input sample.

Multidimensional arrays are also known as tensors. Since tensors is a key concept of the proposed high-order learning model (linear and non-linear), in the following some basic notations and definitions regarding tensor algebra are presented that will be used through out this work. 

\subsection{Tensor Algebra Notations and Definitions}
In this paper tensors are denoted with bold uppercase letters, vectors with bold lowercase letters and scalars with lowercase letters.

\vspace{0.05in}
\noindent \textbf{Tensor vectorization}. The $vec(\bm B)$ operator stacks the entries of a $D$-dimensional tensor $\bm B \in \mathbb R^{p_1 \times \cdots \times p_D}$ on a column vector. 
Specifically, an entry $\bm B=[\cdots b_{i_1, \cdots, i_D}\cdots]$ maps to the $j^{th}$ entry of $vec(\bm B)$, in which $j=1+\sum_{d=1}^D(i_d-1)\prod_{d'=1}^{d-1}p_{d'}$.

\vspace{0.05in}
\noindent \textbf{Tensor products}. Given two tensors $\bm A=[\bm a_1 \cdots \bm a_n] \in \mathbb R^{m \times n}$ and $\bm B=[\bm b_1 \cdots \bm b_q] \in \mathbb R^{p \times q}$ the following products can be defined;
\begin{itemize}
	\item \textit{Kronecker product}. The Kronecker product is the $mp \times nq$ matrix 
	\begin{equation}
	\bm A \otimes \bm B = [\bm a_1 \otimes \bm B \cdots \bm a_n \otimes \bm B] = \begin{bmatrix} a_{11}\bm B & \cdots & a_{1n}\bm B \\ \vdots & \ddots & \vdots \\ a_{m1}\bm B & \cdots & a_{mn}\bm B \end{bmatrix} \nonumber
	\end{equation}
	\item \textit{Khatri-Rao product}. The Khatri-Rao product $\bm A \odot \bm B$ is defined as the 
	$mp \times n$ columnwise Kronecker product, i.e., $\bm A \odot \bm B = [\bm a_1 \otimes \bm b_1 \:\:\: \bm a_2 \otimes \bm b_2\cdots \bm a_n \otimes \bm b_n]$, if $\bm A$ and $\bm B$ have the same number of columns, $n=q$.
	\item \textit{Outer product}. The outer product $\bm b^{(1)} \circ \bm b^{(2)} \circ \cdots \circ \bm b^{(N)}$ of the vectors $\{\bm b^{(i)}\}_{i=1}^N$ forms a tensor whose element at $(i_1, i_2,...,i_N)$ position equals $\prod_{j=1}^N  b_{i_j}^{(j)}$.
\end{itemize}

\vspace{0.05in}
\noindent \textbf{Tensor matricization}.The mode-\textit{d} matricization, $\bm B_{(d)}$, maps a tensor $\bm B$ to a $p_d \times \prod_{d' \neq d}p_{d'}$ matrix by arranging the mode-\textit{d} fibers to be the columns of the resulting matrix. In particular, the $(i_1,\cdots,i_D)$ element of $\bm B$ maps to the $(i_d,j)$ element of $\bm B_{(d)}$, where $j=1+\sum_{d' \neq d}(i_{d'}-1)\prod_{d''<d',d'' \neq d}p_{d''}$.

\vspace{0.05in}
\noindent \textbf{Rank-\textit{R} decomposition}. A tensor $\bm B \in \mathbb R^{p_1 \times \cdots \times p_D}$ admits a rank-\textit{R} decomposition if  $\bm B = \sum_{r=1}^R \bm b_1^{(r)} \circ \cdots \circ \bm b_D^{(r)}$, where the symbol $''\circ''$ represents the vector outer product and $\bm b_d^{(r)}\in \mathbb R^{p_d}$, $d=1,..., D$, $r=1,...,R$. The decomposition can be represented by $\bm B = [\![ \bm B_1, ... ,\bm B_D ]\!]$, where $\bm B_d = [\bm b_d^{(1)}, ... ,\bm b_d^{(R)}] \in \mathbb R^{p_d \times R}$, $d=1,\cdots,D$. When a tensor $\bm B$ admits a rank-\textit{R} decomposition the following results hold;
\begin{equation}
\label{eq:3}
	\bm B_{(d)} = \bm B_d(\bm B_D \odot \cdots \odot \bm B_{d+1} \odot \bm B_{d-1} \odot \cdots \odot \bm B_1)^T
\end{equation}
and 
\begin{equation}
	vec(\bm B) = (\bm B_D \odot \cdots \odot \bm B_1)\bm 1_R,
\end{equation}
where $\bm 1_R$ is a vector of $R$ ones. For more information regarding tensor algebra please refer to \cite{kolda2009tensor}.

\section{High Order Linear Modelling}
\label{sec:high_order_linear}
\subsection{Vector Logistic Regression}
 Let us first assume for simplicity that the input samples are of vector forms. We denote these input vectors as $\bm x \in \mathbb R^{p_1}$, where variable $p_1$ expresses vector dimension.  Using the the logistic regression framework \cite{bishop2006pattern}, one can model the probability function $p^k_w(\cdot)$ as 
 


\begin{equation}
\label{eq:vector_logistic}
	p^k_w(\bm x) = \frac{\exp \bm w^{(k)T} \bm x}{\sum_{i=1}^{C} \exp \bm w^{(i)T} \bm x},
\end{equation}
where $\bm w^{(k)} \in \mathbb R^{p_1}$ with $k=1,2,\cdots, C$ stands for the weights with respect to the $k$-th class. Matrix $\bm W = [\bm w^{(1)} \bm w^{(2)} \cdots \bm w^{(C)}]$ includes all the weights involved in the model. The rational meaning of the weight parameters $\bm w^{(k)}$ is that they express the degree of confidence of the vector input $\bm x$ to belong to the $k$-th out of $C$ available classes. 
In addition, the elements $w^{(k)}_{j}$ of $\bm w^{(k)}=[\cdots w^{(k)}_{j}\cdots]^T$ express the degree of significance of each element of the input vector $\bm x$ with respect to the $k$-th class. 

One simple way to extend Eq. (\ref{eq:vector_logistic}) in case that the inputs are tensors. i.e.,  $\bm X_i \in \mathbb R^{p_1 \times \cdots \times p_D}$, is to take the vectorized forms of them (see the respective text on algebra notation of Section \ref{sec:notation}-{\it tensor vectorization}). The main limitation of such an approach is that (i) a large number of parameters is needed to be estimated, particularly ($C\prod_{l=1}^D p_l$) and (ii) vectorization spoils the spatial structure of tensor inputs, that is, pixels belonging to a neighboring region frequently present similar properties (spatial coherency). A large number of parameters also implies a large number of available labelled observations in order to successfully complete the training procedure. However, usually the available labeled samples are limited due to manual effort required to collect and annotate them.  

\subsection{Matrix Logistic Regression}
\label{sec:matrix_logistic}
In case of matrix input observations, $\bm X \in \mathbb R^{p_1 \times p_2}$, one can reduce the number of model parameters by taking into consideration concepts of \cite{hung2013matrix}, applied for electroencephalogram data classification. Then, the logistic regression model is given by 
\begin{equation}
\label{eq:bilinear}
	p^k_w(\bm X) = \frac{\exp \bm w_1^{(k)T} \bm X \bm w_2^{(k)}}{\sum_{i=1}^{C} \exp \bm w_1^{(i)T} \bm X \bm w_2^{(i)}},
\end{equation}
where $\bm w_1^{(i)} \in \mathbb R^{p_1}$ and $\bm w_2^{(i)} \in \mathbb R^{p_2}$. We also define as $\bm W_1 = [\bm w_1^{(1)} \bm w_1^{(2)} \cdots \bm w_1^{(C)}]$ and $\bm W_2 = [\bm w_2^{(1)} \bm w_2^{(2)} \cdots \bm w_2^{(C)}]$ the aggregate model parameters. 

The key advantage of Eq. (\ref{eq:bilinear}) is that the number of required parameters is  $C(p_1 + p_2)$ instead of $C(p_1 \cdot p_2)$ that would be needed if one vectorize matrix observation input data $\bm X$. This means that the weight parameters $\bm w^{(k)}$, in relation (\ref{eq:bilinear}), for the $k$-th class are {\it rank}-1 canonically decomposed into the weights of $\bm w_1^{(k)} \in \mathbb R^{p_1}$ and $\bm w_2^{(k)} \in \mathbb R^{p_2}$, that is,
\begin{equation}
\label{eq:canonical}
	\bm w^{(k)}=\bm w_2^{(k)} \otimes \bm w_1^{(k)},
\end{equation}
where operator $\otimes$ is the Kronecker product as defined in Section \ref{sec:notation}-{\it Kronecker product}.

Using Eq. (\ref{eq:canonical}), one can write Eq. (\ref{eq:bilinear}) as  
\begin{equation}
\label{eq:bilinear_2}
    p^k_w(\bm X) = \frac{\exp [(\bm w_2^{(k)} \otimes \bm w_1^{(k)})^T vec(\bm X)]}{\sum_{i=1}^{C}\exp [(\bm w_2^{(i)} \otimes \bm w_1^{(i)})^T vec(\bm X)]}.
\end{equation}

\subsection{Tensor-based Logistic Regression}
The aforementioned concept can be extended in case of tensor inputs, $\bm X \in \mathbb R^{p_1\times \cdots \times p_D}$, \cite{zhou2013tensor}. In this way, we have that 

\begin{equation}
\label{eq:tensor_regression}
	p^k_w(\bm X) =\frac{\exp [(\bm w_D^{(k)} \otimes \cdots \otimes \bm w_1^{(k)})^T vec(\bm X)]}{\sum_{i=1}^{C}\exp [(\bm w_D^{(i)} \otimes \cdots \otimes \bm w_1^{(i)})^T vec(\bm X)]},  
\end{equation} 
where $\bm w_l^{(k)} \in \mathbb R^{p_l}$ with $l=1,2,\cdots, D $ is the $l$-th {\it rank}-1 canonical decomposition of the weight parameters $\bm w^{(k)}$ for the $k$-th class. This property is referred as CANDECOMP, stem from  CANonical DECOMPosition, or PARAFAC, stem from PARAllel FACtors,  in the literature \cite{harshman1970foundations, carroll1970analysis}. That is,  

\begin{equation}
\label{eq:canonical2}
	\bm w^{(k)}=\bm w_D^{(k)} \otimes \cdots \otimes \bm w_1^{(k)}.
\end{equation}

Eq. (\ref{eq:tensor_regression}) can be seen as an expression of the Khatri-Rao product, denoted as operator $\odot$ (see Section \ref{sec:notation}), which is the column-wise Kronecker product of the {\it rank}-1 canonical decomposition weight parameters $\bm w_l^{(k)}$, 

\begin{equation}
\label{eq:tensor_regression2}
	p^k_w(\bm X) =\frac{\exp [(\bm w_D^{(k)} \odot \cdots \odot \bm w_1^{(k)})^T vec(\bm X)]}{\sum_{i=1}^{C}\exp [(\bm w_D^{(i)} \odot \cdots \odot \bm w_1^{(i)})^T vec(\bm X)]},
\end{equation} 

In Eqs. (\ref{eq:tensor_regression}) and (\ref{eq:tensor_regression2}), we can aggregate the total weight parameters as 
\begin{equation}
\label{eq:aggregate_weights}
\bm W_l = [\bm w_l^{(1)} \bm w_l^{(2)} \cdots \bm w_l^{(C)}],
\end{equation} 
with $l=1,2,\cdots,D$. In other words, matrix $\bm W_l$ contains all the weight parameters with respect to the $l$-th dimension of the tensor $\bm X$ overall classes, i.e, related with the $p_l$ tensor dimension.

The representation of Eqs. (\ref{eq:tensor_regression}) and (\ref{eq:tensor_regression2}) significantly reduces the number of model parameters needed for classifying the tensor inputs $\bm X$. In particular, the vector-based logistic regression model derived through vectorization of tensor $\bm X$ requires the estimation of $C\prod_{l=1}^D p_l$ parameters, while the tensor-based representation  of (\ref{eq:tensor_regression}) and (\ref{eq:tensor_regression2}) reduces this number to $C\sum_{l=1}^D p_l$. 

The advantages of the aforementioned proposed representation for the classification of hyperspectral images are twofold. First, as we reduce the number of weight parameters, a smaller number of labelled data samples is required to train the logistic regression model. This is an important factor for developing classifiers that can better generalize to unseen hyperspectral data. 
Usually, the manual effort for the annotation is laborious and therefore a small number of labelled training data is available. Second, the {\it rank}-1 canonical decomposition of the model weights provides a physical interpretation of how the spatial and spectral information of the hyperspectral tensor input $\bm X$ contributes to the classification. In particular, according to the statements made right after Eq. (\ref{eq:vector_logistic}), the weights $\bm w^{(k)}$ express the degree of importance of the tensor input $\bm X$ to belong to the $k$-th class. Since these weights are canonically decomposed into $D$ separate weights $\bm w_l^{(k)}$, each indicating the contribution of $p_l$ tensor dimension to the belong to the $k$-th class, the proposed model provides a quantitative representation of how the elements of each tensor dimension tunes
the classification performance. 

More specifically, in case of hyperspectral imaging, the dimension of tensor inputs equals 3, i.e., $D=3$. The first two dimensions refer to the spatial properties of the pixel data, while the third one to the spectral bands. Therefore, the first two decomposed weights, i.e., $\bm w^{(1)}$ and $\bm w^{(2)}$, express how the pixel spatial coherency affects the classification outcome. On the other hand, the third decomposed weight vector $\bm w^{(3)}$ indicates how the spectral bands influence the classification and which of the spectral bands are the most salient. This property of the proposed model is very important towards the analysis of hyperspectral data since it facilitates the interpretation of the results and quantifies the importance of the spectral bands on the classification compressing the influence of the bands that are of less importance. 

\subsection{Estimation of the Model Weights}
\label{sec:estimation}
The model weight parameters are estimated through a training set $\mathcal S = \{(\bm X_i, \bm t_i)\}_{i=1}^N$ of $N$ samples as Eq. (\ref{eq:dataset}) indicates. We recall that if $\bm X_i$ belongs to the $k^{th}$ class, then $\bm t_i$ is a vector with all elements zero except the element $k$, which equals one, i.e., $t_{i,j}=0,\forall j\neq k$ and $t_{i,k}=1$. 
By minimizing the negative log-likelihood function,
\begin{equation}
\label{eq:neg_log_likelihood}
    L(\bm W_1,...,\bm W_D ;\mathcal S) = -\sum_{i=1}^N \sum_{k=1}^{C} t_{i,k}\log p^k_w(\bm X_i),
\end{equation}
the weight parameters can be estimated as
\begin{equation}
\label{eq:negative_log_likelihood_t}
    \{ \bm{\hat W_1}, \bm{\hat W_2}, \cdots, \bm{\hat W_D} \} =\arg\min_{\forall \bm W_i}L(\bm W_1,...,\bm W_D ;\mathcal S). 
\end{equation}

 In Eq. (\ref{eq:negative_log_likelihood_t}), matrices $\bm{ \hat W_l}, l=1,\cdots,D$ refer to the optimal estimates of the $\bm W_l$ weight parameters, expressing the contribution of the $l$-dimension of the tensor input to the classification for all $C$ available classes.

Based on the statements of Section \ref{sec:notation}, presenting some basic notations on tensor algebra ({\it Rank-R decomposition and tensor matricization}), it is held that   

\begin{equation}
\label{eq:observation1}
\begin{split}
	&\langle \bm w_D^{(k)} \odot \cdots \odot \bm w_1^{(k)}, \bm X \rangle = \\
	 &\langle \bm w_l^{(k)}, \bm X_{(l)}(\bm w_D^{(k)}\odot \cdots \odot \bm w_{l+1}^{(k)}\odot \bm w_{l-1}^{(k)}\odot \cdots \odot \bm w_1^{(k)}) \rangle.
\end{split}
\end{equation}

The proof of Eq. (\ref{eq:observation1}) is given in Appendix A. In Eq. (\ref{eq:observation1}), $\bm X_{(l)}$ denotes the mode-$l$ matricization of tensor $\bm X$ obtained by keeping the $l$-dimension intact and concatenating the slices of the rest dimensions into a long matrix \cite{kolda2009tensor}. We also recall that the operator $\odot$ refers to the Khatri-Rao product, while the $\langle \cdot,\cdot\rangle$ to the inner product between two tensors. 

The left hand of Eq. (\ref{eq:observation1}) expresses the arguments of $\exp(\cdot)$ involved in the linear logistic regression model of Eq. (\ref{eq:tensor_regression2}). Therefore, (\ref{eq:tensor_regression2}) can be rewritten by taking into account the right-hand of (\ref{eq:observation1}). In the sequel, generation of (\ref{eq:observation1}) over all available classes is obtained as 

\begin{equation}
\label{eq:observation_general}
\begin{split}
	&\langle \bm W_D \odot \cdots \odot \bm W_1, \bm X \rangle = \\
	 &diag(\langle \bm W_l, \bm X_{(l)}(\bm W_D \odot \cdots \odot \bm W_{l+1}\odot \bm W_{l-1}\odot \cdots \odot \bm W_1) \rangle).
\end{split}
\end{equation}

As we can see Eq. (\ref{eq:observation1}) is actually the inner product of two vectors. The first is the weight parameters $\bm w_l^{(k)}$ expressing the contribution of the $l$ dimension of the tensor input as far as the $k$-th class is concerned. On the other hand, the second vector is independent from the values of $\bm w_l^{(k)}$. Therefore, one approach for optimally estimating the weights $\bm w_l^{(k)}$ and consequently $\bm w^{(k)}$ [see Eq. (\ref{eq:canonical2})] is one to consider all the weight parameters  $\bm w_q^{(k)}$ with $q\neq l$ apart from the $l^{th}$ fixed and then solving with respect to $\bm w_l^{(k)}$. This procedure is iteratively applied for all weight parameters $\bm w_l^{(k)}$ until some termination criteria are met. Similar statements can be concluded for the matrix representation of Eq. (\ref{eq:observation_general}). 

Actually, the learning algorithm used to optimally estimate the model weights in case that Eq. (\ref{eq:canonical2}) is held, i.e,. the {\it rank}-1 canonical decomposition, simulates a regression learning where we use as inputs the data of the right-hand of Eq. (\ref{eq:observation_general}), that is,  
\begin{equation}
\label{eq:transformed}
	\bm X_{(l)}(\bm W_D \odot \cdots \odot \bm W_{l+1} \odot \bm W_{l-1} \odot \cdots \odot \bm W_1).
\end{equation}
Therefore, the parameter estimation problem can be solved by using conventional software such as scikit-learn in python\footnote{http://scikit-learn.org/}. The proposed learning procedure, considering the {\it rank}-1 canonical decomposition of the weight parameters is described in Algorithm \ref{alg:1}. 

\begin{figure}[!h]
{\begingroup
\removelatexerror
\begin{algorithm}[H]
\caption{Optimal Estimation of Model Weights using {\it rank}-1 Canonical Decomposition}
\label{alg:1}
\SetAlgoLined
 \textbf{Initialization:} \\
 1. Set Iteration Index $n\rightarrow 0$\\
 2. Randomize all the weight parameters $\bm W_l(n) \in \mathbb R^{p_l \times C}$ \\
 for all $l=1,...,D$ \\
 3. \Repeat{termination criteria are met}{
  \For{$l=1,...,D$} {
   3.1 Calculate the matrix of Eq.(\ref{eq:transformed}), that is, \\
   $\bm X_{(l)}(\bm W_D(n) \odot \cdots \odot \bm W_{l+1}(n) \odot \bm W_{l-1}(n) \odot \cdots \odot \bm W_1(n))$\\
   3.2 Update matrix $\bm W_l(n) $ using a regression learning algorithm through minimization of Eq. (\ref{eq:negative_log_likelihood_t}) such as \\
   $\bm W_l(n+1) \rightarrow$
   $\arg \min\limits_{\bm W_l} L(\bm W_1(n+1),...,$
   $\bm W_{l-1}(n+1),\bm W_{l}(n),...,\bm W_D(n))$
  	}
   Set $n \rightarrow n+1$
  }
\end{algorithm}
\endgroup}
\end{figure}

Although, the high-order linear model can provide physically interpretable results, due to its structure, it is restricted to produce decision boundaries that are linear in the input space. This implies that this model is not able to cope with more complicated problems, where non-linear decision boundaries are necessary to provide classification results of high accuracy. This motivates us to extend the previous linear regression model to a non-linear one. The proposed non-linear model should assume again a {\it rank}-1 canonical decomposition of the weight parameters in order to retain the aforementioned advantages in hyperspectral classification.

\section{High-Order Non-linear Modelling}
\label{sec:nonlinear}
The proposed high-order non-linear model is based on the concepts of the previously discussed linear logistic regression filters with the difference that a nonlinear transformation is applied on the input data. This means, in other words, that the probability $p^k_w(\cdot)$ of an input observation  $\bm X$ to belong to one of the $C$ available classes is non-linearly interwoven with respect to the input tensor data and the weight parameters that influence the importance of these data on classification performance through a function $f_w(\cdot)$, i.e.,

\begin{equation}
\label{eq:nonlinear}
      p^k_w(\cdot)=f_w(\bm X).
\end{equation}

The main difficulty in implementing Eq. (\ref{eq:nonlinear}) is that function $f_w(\cdot)$ is actually unknown. One way to parameterize the unknown function $f_w(\cdot)$ is to exploit the principles of the universal approximation theorem, stating that a function, under some assumptions of continuity, can be approximated by a feedforward neural network with a finite number of neurons within any degree of accuracy \cite {ndoulam}. Feedforward neural networks have been proven as a reliable framework for image classification \cite{adoulam1}. 

However, the main difficulty in applying a feedforward neural network (FNN) for hyperspectral classification problems is twofold. First, a large number of weight parameters is required to be learned, proportionally to $Q\prod_{l=1}^D p_l + QC$, where variable $Q$ refers to the number of hidden neurons of the network. This outcome derives as a consequence of the structure of the network as is briefly described below (see Section \ref{sec:FNN}). This, in the sequel, implies that a large number of labelled samples needed to successfully train the network. Second, the weights of the network are not directly related to the physical properties of hyperspectral data and how these properties affect the classification performance, since the inputs are vectorized and thus they do not preserve their structure. 
Networks are often treated as "black boxes" when they are applied for classification of hyperspectral data. To overcome these problems, we propose a modification of conventional feedforward neural network structures so that network weights from the input to the hidden layer satisfy the {\it rank}-1 canonical decomposition according to statements of the previous section. In addition, we introduce a learning algorithm to train the so-called {\it rank}-1 canonical decomposition feedforward neural network - {\it rank}-1 FNN. Before proposing {\it rank}-1 FNN, we briefly describe how $p^k_w(\cdot)$ is modelled through a FNN.   

\subsection{FeedForward Neural Network Modelling}
\label{sec:FNN}

\begin{figure*}[t]
  \begin{minipage}{1.0\linewidth}
    \centering
    \centerline{\includegraphics[width=0.7\linewidth]{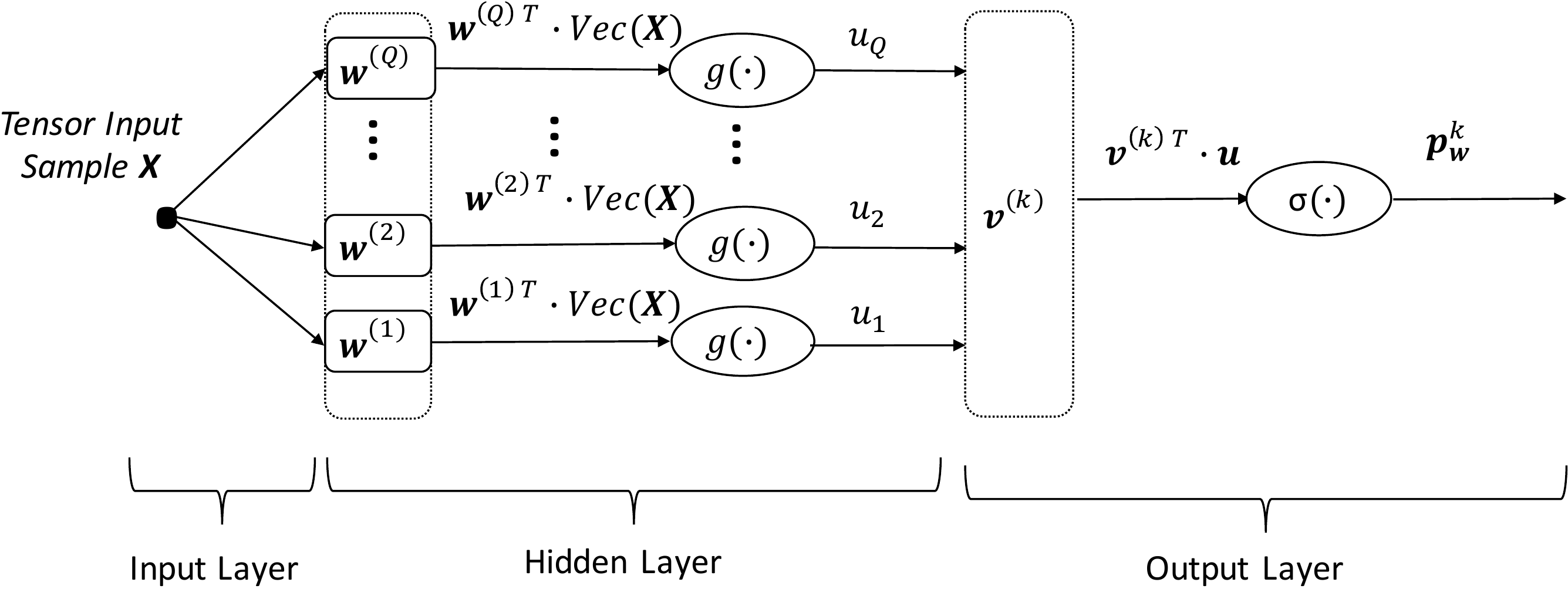}}
  \end{minipage} 
  \caption{The structure of a feedforward neural network.}
  \label{fig:FNN}
\end{figure*}

Fig. \ref{fig:FNN} illustrates a feedforward neural network nonlinearly approximating the probability $p^k_w(\cdot)$. Initially, the tensor input $\bm X$ is vectorized creating a vector, $vec(\bm X)$, of size $\prod_{l=1}^D p_l$. The network is assumed to have $Q$ hidden neurons. Each neuron is associated with an activation function $g(\cdot)$. In this paper, the sigmoid function $g(x)=1/(1+\exp(-x))$ is selected, where factor $a$ regulates the steepness of the function. The activation function of the $i$-th out of $Q$ hidden neurons receives as input the inner product of $vec(\bm X)$ and a weight vector $\bm w^{(i)}$ 
associated with the $i$-th neuron and produces as output a scalar $u_i$ given by \cite{adoulam}
\begin{equation}
\label{eq:basic_model_in}
	u_i = g(\bm w^{(i)T} vec(X))\equiv g(\langle 	\bm w^{(i)}, \bm vec(X) \rangle),
\end{equation}
where we recall that the operator $\langle \cdot,\cdot \rangle $ expresses the inner product. It should be mentioned that in the current notation the superscript $i$ of the weights $\bm w^{(i)}$ refers to the $i$-th hidden neuron. Gathering the responses of all hidden neurons in one vector $\bm u=[u_1, u_2,\cdots,u_Q]^T$, we have that
\begin{equation}
\label{eq:basic_model_in}
	\bm u = g(\langle \bm W, \bm X \rangle),
\end{equation}
where $\bm W=[\bm w^{(1)},\cdots, \bm w^{(Q)}]^T$ is a matrix containing the weights $\bm w^{(i)}$ for all hidden neurons, $i=1,\cdots, Q$. 
Thus, the output of the network is given as  
\begin{equation}
\label{eq:basic_model}
	\bm p^{k}_w =\sigma ( \langle \bm v^{(k)},\bm u \rangle)\equiv \sigma (\bm v^{(k)T} \bm u), 
\end{equation}
where $\sigma(\cdot)$ stands for the softmax function, $\bm v^{(k)} $ the weights between the hidden and the output layer and the superscript of the weights for the $k$-th class. The softmax function corresponds to the following conditional probability and is defined as follows for a class $i$

\begin{equation}
\label{eq:sofmax}
	P(y=i|x)=\frac{\exp( {\bm x^T} \cdot {\bm w}_i) }{\sum_{j=1}^{j=C}{\exp( {\bm x^T} \cdot {\bm w}_j)}}.
\end{equation}

\subsection{{\it Rank}-1 Canonical Decomposition Feedforward Neural Networks - {\it Rank}-1 FNN}
To reduce the number of parameters of the network and to relate the classification results to the spatial and spectral properties of hyperspectral input data, we {\it rank}-1 canonically decomposed the weight parameters $\bm w^{(i)}$ as in Eq. (\ref{eq:canonical2}).  
We should stress that vector $\bm w_l^{(i)}$ relates the input tensor data with the $i$-th hidden neuron and therefore $i=1,2,...,Q$.

Then, taking into account the properties of Eq. (\ref{eq:observation1}), the output of the $i$-th hidden neuron $u_i$ can be written as 
\begin{equation}
\label{eq:basic_model_in_2}
\begin{split}
	u_i & = g(\langle \bm w^{(i)}, \bm X \rangle)\\
	&=g(\langle \bm w_D^{(i)} \otimes \cdots \otimes \bm w_1^{(i)}, \bm X \rangle) \\
	&= g(\langle \bm w_D^{(i)} \odot \cdots \odot \bm w_1^{(i)}, \bm X \rangle) \\
	&= g(\langle \bm w_l^{(i)}, \bm \tau_{\neq l}^{(i)} \rangle).
\end{split}
\end{equation} 
In Eq. (\ref{eq:basic_model_in_2}), vector $\bm \tau_{\neq l}^{(i)}$ is a transformed version of tensor input $\bm X$. Vector $\bm \tau_{\neq l}^{(i)}$ is independent from $\bm w_l^{(i)}$.That is, 
\begin{equation}
\label{eq:tau}
\tau_{\neq l}^{(i)}= \bm X_{(l)}(\bm w_D^{(i)}\odot \cdots \odot \bm w_{l+1}^{(i)}\odot \bm w_{l-1}^{(i)}\odot \cdots \odot \bm w_1^{(i)}),
\end{equation} 
where we recall that $\bm X_{(l)}$ is the mode-$l$ matricization of $\bm X$. 

\begin{figure*}[t]
  \begin{minipage}{1.0\linewidth}
    \centering
    \centerline{\includegraphics[width=0.9\linewidth]{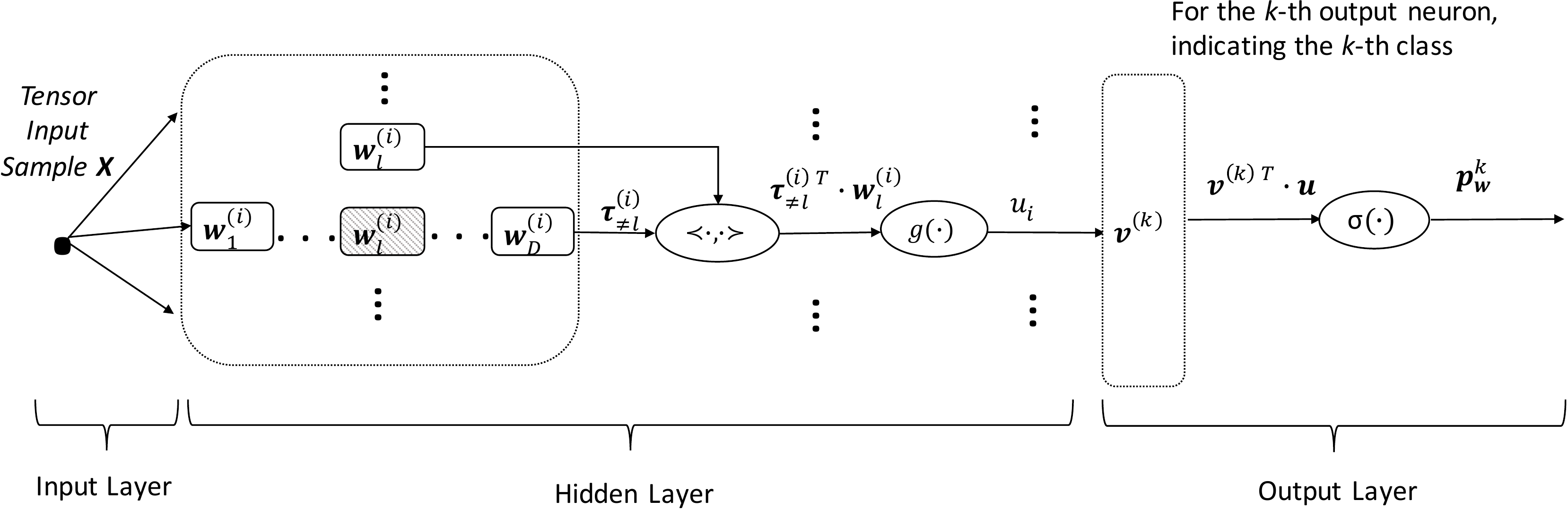}}
  \end{minipage} 
  \caption{The structure of the proposed {\it rank}-1 feedforward neural network.}
  \label{fig:R1_FNN}
\end{figure*}

Eq. (\ref{eq:basic_model_in_2}) actually resembles the operation of a single perceptron with inputs the weights $\bm w_l^{(i)}$ and the transformed version $\tau_{\neq l}$ of the input data. That is, if the {\it rank}-1 canonically decomposed weights $\bm w_r^{(i)}$ with $r\neq l $ are known then $\tau_{\neq l}^{(i)}$ will be also known. Fig. \ref{fig:R1_FNN} shows the structure of the proposed {\it rank}-1 FNN. The model consists of one input layer, one hidden and one output layer. The main modification of this structure compared to a conventional FNN is in the hidden layer. More specifically, the weights of a hidden neuron are first decomposed into $D$ canonical factors, each expresses the spatial and spectral band contribution to the classification performance. In this figure, we have shaded the weight vector $\bm w_l^{(i)}$ since all the rest weight vectors are used to estimate the transformed input $\tau_{\neq l}^{(i)}$.

\subsection{The Learning Algorithm}
\label{sec:learning_nonlinear}
To train the proposed {\it rank}-1 FNN model the set $\mathcal S = \{ (\bm X_i, \bm t_i) \}_{i=1}^N$ containing $N$ samples is used. The learning algorithm minimizes the negative log-likelihood (see relation (\ref{eq:neg_log_likelihood})) with respect to network responses $\bm y_i=[\cdots y_{i,k}\cdots]^T$, with $y_{i,k}\equiv p_w^k(\bm X_i)$, and targets $\bm t_i$ over all training samples.


In case of using a conventional neural network training algorithm, the estimated weights do not satisfy the {\it rank}-1 canonical decomposition assumption expressed by Eq. (\ref{eq:canonical2}). For this reason, in the proposed learning algorithm we initially fix all the weights  $\bm w_r^{(i)}$ with $r\neq l$. This way, we are able to estimate the transformed input vector $\bm \tau_{\neq l}^{(i)}$. Therefore, the only unknown parameters of the hidden layer is the vector $\bm w_l^{(i)}$. This can be derived through a gradient based optimization algorithm, assuming that the derivative $\partial L / \partial \bm w_l^{(i)}$ is known. Then, the weights are updated towards the negative direction of the partial derivative. 

One way to estimate the partial derivative $\partial L / \partial \bm w_l^{(i)}$  is to exploit principles of the back-propagation algorithm which actually implements the chain rule property for estimating the derivative of the error with respect to the network weights. In particular, by using the back-propagation algorithm, we compute the partial derivatives
\begin{equation}
\label{eq:partial}
\begin{split}
    &\partial L / \partial \bm w_l^{(i)} \text{ for } l=1,...,D \text{ and } i=1,...,Q \\
    &\partial L / \partial \bm v^{(k)} \text{ for } k=1,...,C.
\end{split}
\end{equation}
The partial derivative $\partial L / \partial \bm w_l^{(i)}$ can be estimated if we assume all the weights $\bm w_r^{(i)}$ with $r\neq l $ fixed, since in this case we can estimate the vector $\bm \tau_{\neq l}^{(i)}$. Therefore, estimation of the parameters of the {\it Rank}-1 FNN is obtained by iteratively solving with respect to one of the $D$ canonical decomposed weight vectors, assuming the remaining fixed. Algorithm \ref{alg:2} presents the main steps of the proposed algorithm, applying for the calculation of the hidden layer weights of the network. 

\begin{figure}[t]
{\begingroup
\removelatexerror
\begin{algorithm}[H]
\caption{Optimal Estimation of Hidden Layer Model Weights of the {\it Rank}-1 FNN}
\label{alg:2}
\SetAlgoLined
 \textbf{Initialization:}\\
 1. Set Iteration Index $n\rightarrow 0$\\
 2. Randomize all the weight vectors $\bm w_l^{(i)}(n) \in \mathbb R^{p_l \times 1}$ \\
 for all $l=1,...,D$ and for all $i=1,2,\cdots, Q$ \\
 3. Randomize all the weight vectors $\bm v^{(k)}(n) \in \mathbb R^{Q \times 1}$ \\
 for all $k=1,...,C$ \\
 4. \Repeat{termination criteria are met}
                 {
  \For{$l=1,...,D$} {
            \For {$i=1,...Q$}
              {
                 4.1 Estimate the transformed input vector $\bm \tau_{\neq l}^{(i)}$ \\
                 $\tau_{\neq l}^{(i)}= \bm X_{(l)}(\bm w_D^{(i)}(n)\odot \cdots \odot \bm w_{l+1}^{(i)}(n)\odot \bm w_{l-1}^{(i)}(n+1)\odot \cdots \odot \bm w_1^{(i)}(n+1)),$ \\
                 4.2 Estimate the derivative error of  $\partial L / \partial \bm w_l^{(i)}$ \\
                 4.3 Update the weights $\bm w_l^{(i)}(n)$ towards the negative direction of derivative 
              }
   
  	}
  	\For{$k=1,...,C$} {
  	4.4 Estimate the derivative error of  $\partial E / \partial \bm v^{(k)}$ \\
    4.5 Update the weights $\bm v^{(k)}(n)$ towards the negative direction of derivative  
  	}
  	Set $n \rightarrow n+1$
  }
\end{algorithm}
\endgroup}
\end{figure}

\begin{figure*}[t]
  \begin{minipage}{0.5\linewidth}
    \centering
    \centerline{\includegraphics[height=5.5cm]{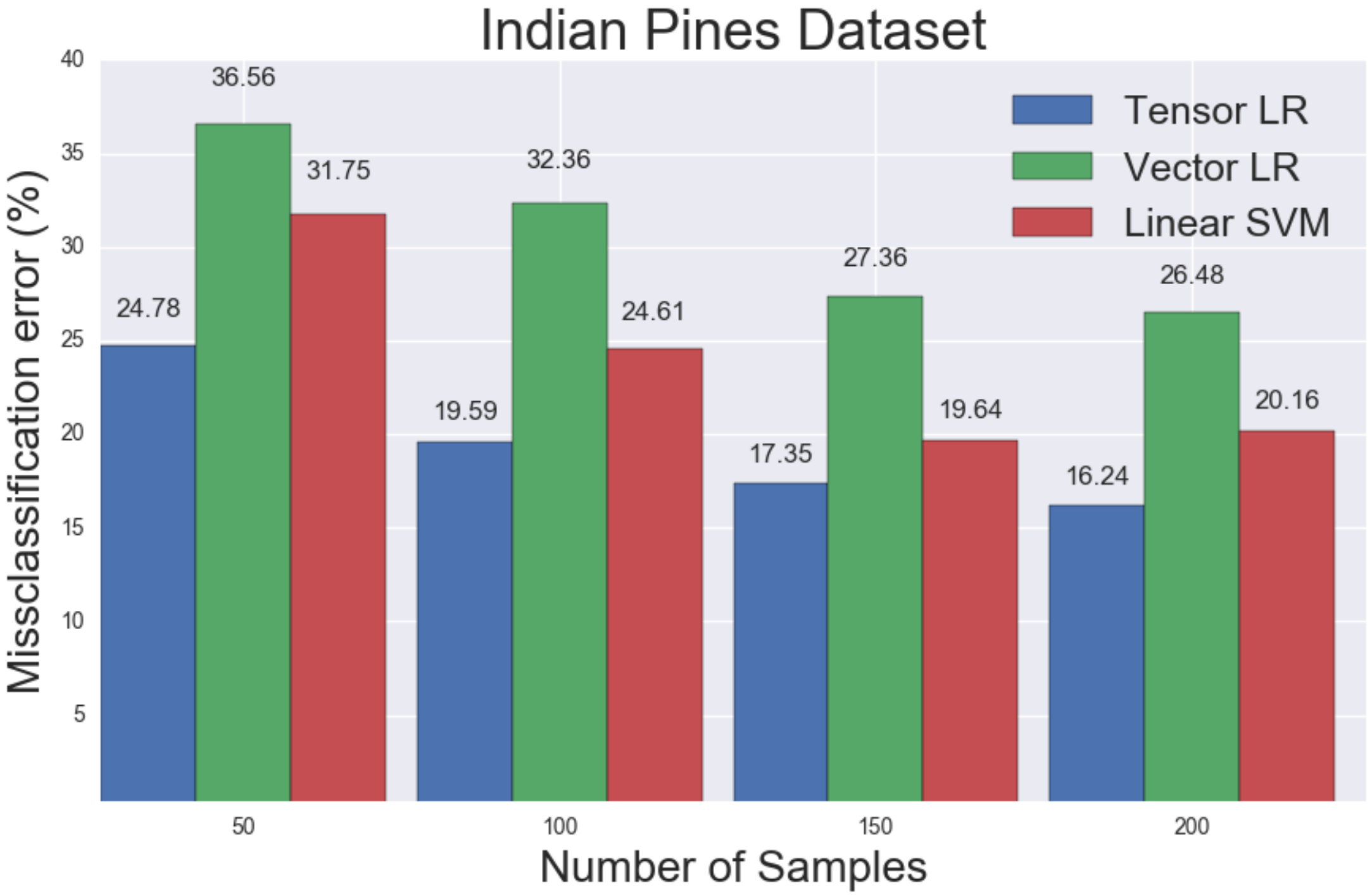}}
  \end{minipage} 
  \begin{minipage}{0.5\linewidth}
    \centering
    \centerline{\includegraphics[height=5.5cm]{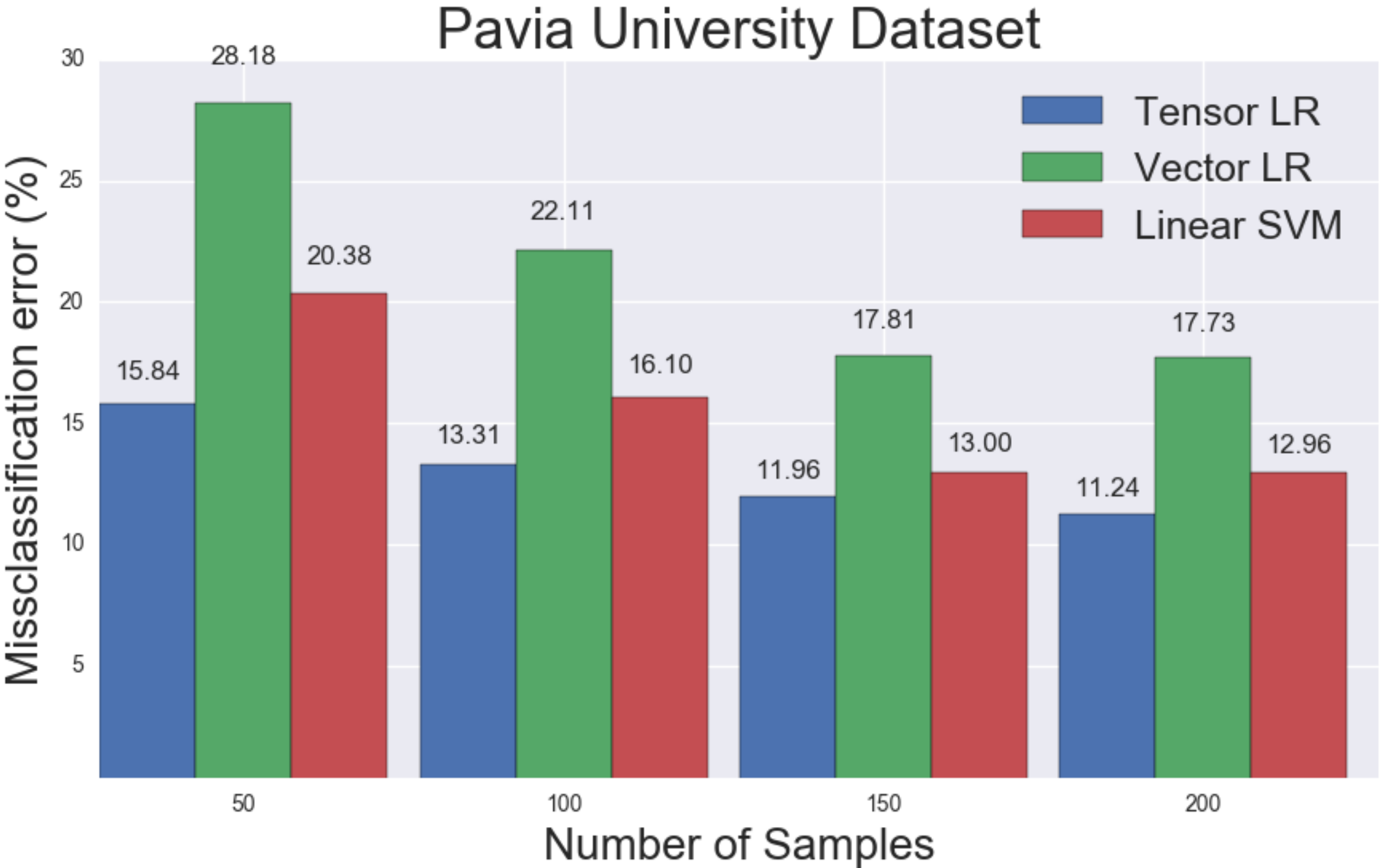}}
  \end{minipage}
  \caption{Misclassification error on test set for tensor-based logistic regression, vector-based logistic regression and linear SVMs.}
  \label{fig:1}
\end{figure*}

\section{EXPERIMENTAL RESULTS ON HYPERSPECTRAL DATA}
\label{sec:experiments}
A natural question that arises is whether such a reduction in the number of parameters would limit the descriptive power of the models in (\ref{eq:tensor_regression}) and in (\ref{eq:basic_model}). To answer this question we present quantitative results regarding classification accuracy. Furthermore, we compare the performance of the linear high-order model (tensor-based logistic regression) against two other well-known linear models; vector-based logistic regression and linear SVMs, and the performance of the high-order nonlinear model against Fully Connected Feed Forward Neural Nets (FCFFNN), nonlinear SVMs and two deep learning approaches for classifying hyperspectral data; an approach based on Stacked-Autoencoders \cite{chen2014deep} and an approach based on the exploitation of Convolutional Neural Networks \cite{Makantasis-etal:15}. These methods are the current state of the art in the literature. 

The model in (\ref{eq:tensor_regression}) is linear in the feature space and the estimated parameters can be used to estimate the importance of the spectral bands. In particular, the most important feature elements should have the highest, in absolute value, weight coefficients, while feature elements uncorrelated with the output variables should have coefficient values close to zero. This way, we can reduce the dimensionality of raw data keeping feature elements with the highest weight coefficients.

\subsection{Datasets}
In our study we used AVIRIS and ROSIS sensors hyperspectral datasets \footnote{The code for the Rank-1 FNN is available at https://bit.ly/2GRMd85}. In particular, we used i) the Indian Pines dataset, which depicts a test site in North-western Indiana and consists of 224 spectral reflectance bands and ii) the Pavia university datasets, whose number of spectral bands are 103.

A hyperspectral image is represented as a 3D tensor of dimensions $p_1 \times p_1 \times p_3$, where $p_1$ and $p_2$ correspond to the height and width of the image and $p_3$ to the spectral bands. In order to classify a pixel $I_{x,y}$ at location $(x,y)$ on image plane and successfully fuse spectral and spatial information, we use a square patch of size $s \times s$ centered at pixel $I_{x,y}$. Let us denote as ${\bm t}_{x,y}$ the class label of the pixel at location $(x,y)$ and as ${\bm X}_{x,y}$ the tensor patch centered at pixel $I_{x,y}$. Then, we can form a dataset $S=\{({\bm X}_{x,y}, {\bm t}_{x,y}) \}$ for $x=1,2,\cdots,p_2$ and $y=1,2,\cdots,p_1$. Each one of the patches ${\bm X}_{x,y}$ is also a 3D tensor with dimension $s \times s \times p_3$, which contains spectral and spatial information for the pixel located at $(x,y)$. The dataset $S$ is used to train the classifiers. The Pavia University and the Indian Pines datasets contain $42,776$ and $10,086$ labeled pixels respectively.

\subsection{Classification Accuracy of the Tensor-based Logistic Regression}
\label{sub:tensor_based_logistic_regression}
 The performance of the tensor-based logistic regression method is evaluated into four different experiments, each of different number of training samples to assess classification accuracy even in cases where a small number of samples is used. The evaluation is compared against vector-based logistic regression and linear SVMs. For each of the four experiments, we use as training dataset a subset of $S$ that contains $50$, $100$, $150$ and $200$ samples from each class respectively. The remaining samples are used as test data. 

The same training data are used to train all models. The vector-based logistic regression as well as the linear SVMs are trained by using the vectorized versions of patches ${\bm X}_{x,y}$ that belong to the training datasets. The tensor-based logistic regression model is trained by using the same patches without applying any transformation on them, and thus, keeps intact their spatial structure. 

Figure \ref{fig:1} presents the classification accuracy, on the test set, for all tested methods. In both datasets and in all cases the tensor-based model outperforms linear SVMs and vector-based logistic regression, despite the fact that it employs the smallest number of parameters. These results quantitatively answer the question regarding the capacity of the model in (\ref{eq:tensor_regression}) under a small sample setting framework.

\subsection{Tensor-based Dimensionality Reduction}
In the following we evaluate the quality of the dimensionality reduction that can be conducted by the tensor-based logistic regression model. Towards this direction, we utilize the model presented in \cite{Makantasis-etal:15}, where a deep Convolutional Neural Network (CNN) is used for classifying hyperspectral data. The authors reduce the dimensionality of the data along the spectral dimension by applying principal components analysis and by utilizing the $n$ principal components that preserve at least $99.9\%$ of the total dataset variance. 

In this work we utilize the same CNN structure, but we reduce the dimensionality of the raw data by selecting the $n$ most significant spectral bands, i.e., spectral bands to which the tensor-based logistic regression model assigns the larger, in terms of absolute value, coefficients. Due to the fact that this method does not take the variation of estimation into account, we firstly normalized the data, so as to suppress the effect of variance. The results in terms of classification accuracy on the test set, using the same CNN for both dimensionality reduction methods are presented in Table \ref{table:pavia_uni}. Each experiment has been executed at 10 different runs and the standard deviation across all different runs is also depicted in Table \ref{table:pavia_uni}. In this table, we denote as TB-CNN the tensor-based dimensionality reduction followed by a CNN classifier and as PCA-CNN the PCA dimensionality reduction followed again by a CNN model. We conducted three experiments where we use $10\%$, $20\%$ and $40\%$ of the datasets as the training samples.

\begin{table}[t]
\centering
\caption{Overall classification accuracy results (\%) of the tensor-based logistic regression model for both datasets.}
\newcolumntype{L}[1]{>{\hsize=#1\hsize\raggedright\arraybackslash}X}%
\newcolumntype{C}[1]{>{\hsize=#1\hsize\centering\arraybackslash}X}%
\label{table:pavia_uni}

\begin{tabularx}{0.98\linewidth}{L{7}C{6}C{6}C{6}}
\hline \hline \\
\multicolumn{4}{c}{{\bf Pavia University}} \\ \hline 

Splitting ratio & 10\% & 20\% & 40\% \\ \hline

TB-CNN  & 97.33 $\pm$ 0.5 & 98.41 $\pm$ 0.3 & 99.31 $\pm$ 0.2    \\ \hline

PCA-CNN & 97.58 $\pm$ 0.3 & 98.53 $\pm$ 0.3 & 99.42 $\pm$ 0.1    \\ \hline \hline \\

\multicolumn{4}{c}{{\bf Indian Pines}}  \\ \hline 

Splitting ratio & 10\% & 20\% & 40\%   \\ \hline 

TB-CNN  & 82.29  $\pm$ 1.1 & 90.09 $\pm$ 0.7 & 95.98 $\pm$ 0.6  \\ \hline

PCA-CNN & 84.34 $\pm$ 0.9 & 91.72 $\pm$ 0.7 & 96.37 $\pm$ 0.5  \\ \hline \hline

\end{tabularx}
\end{table}

\begin{figure*}[t]
  \begin{minipage}{0.5\linewidth}
    \centering
    \centerline{\includegraphics[width=1.0\linewidth]{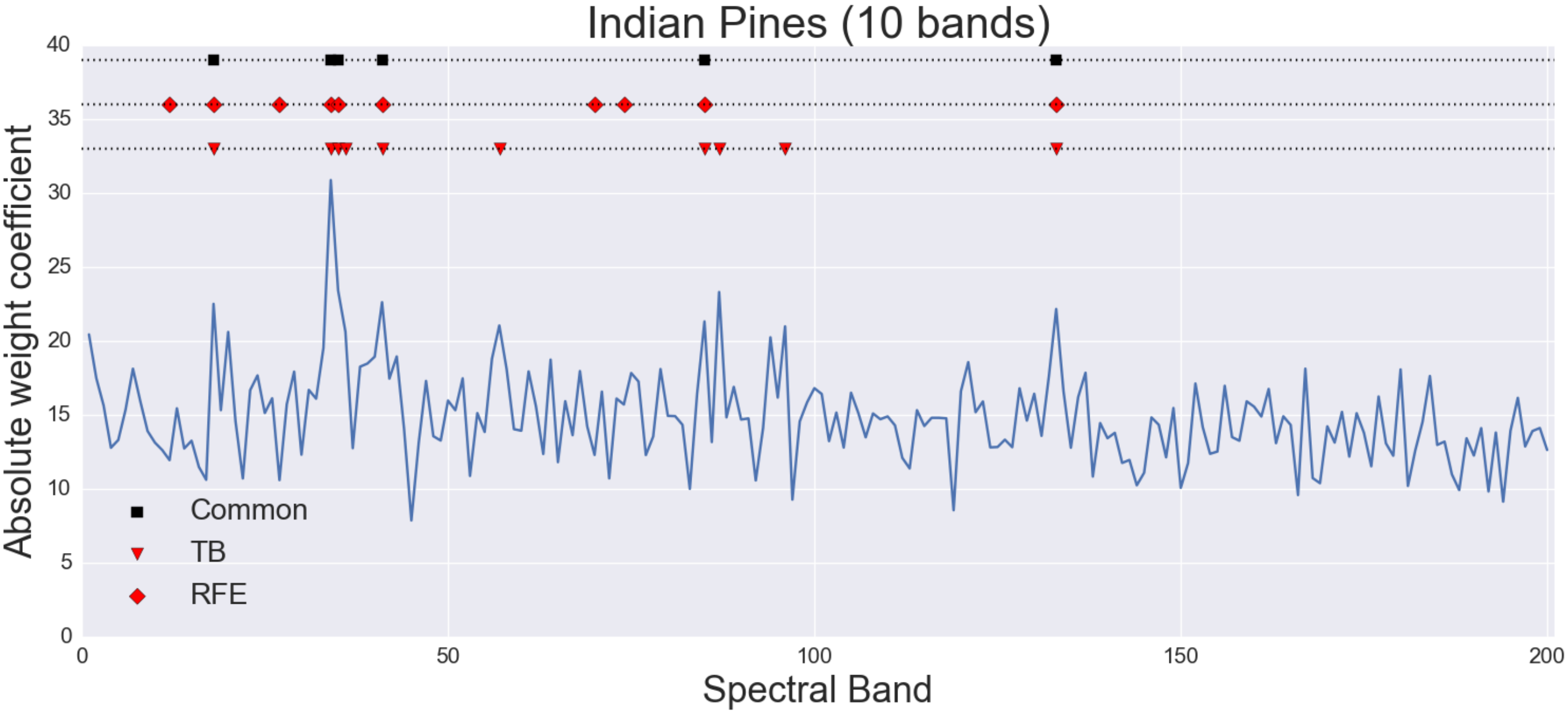}}
  \end{minipage} 
  \begin{minipage}{0.5\linewidth}
    \centering
    \centerline{\includegraphics[width=1.0\linewidth]{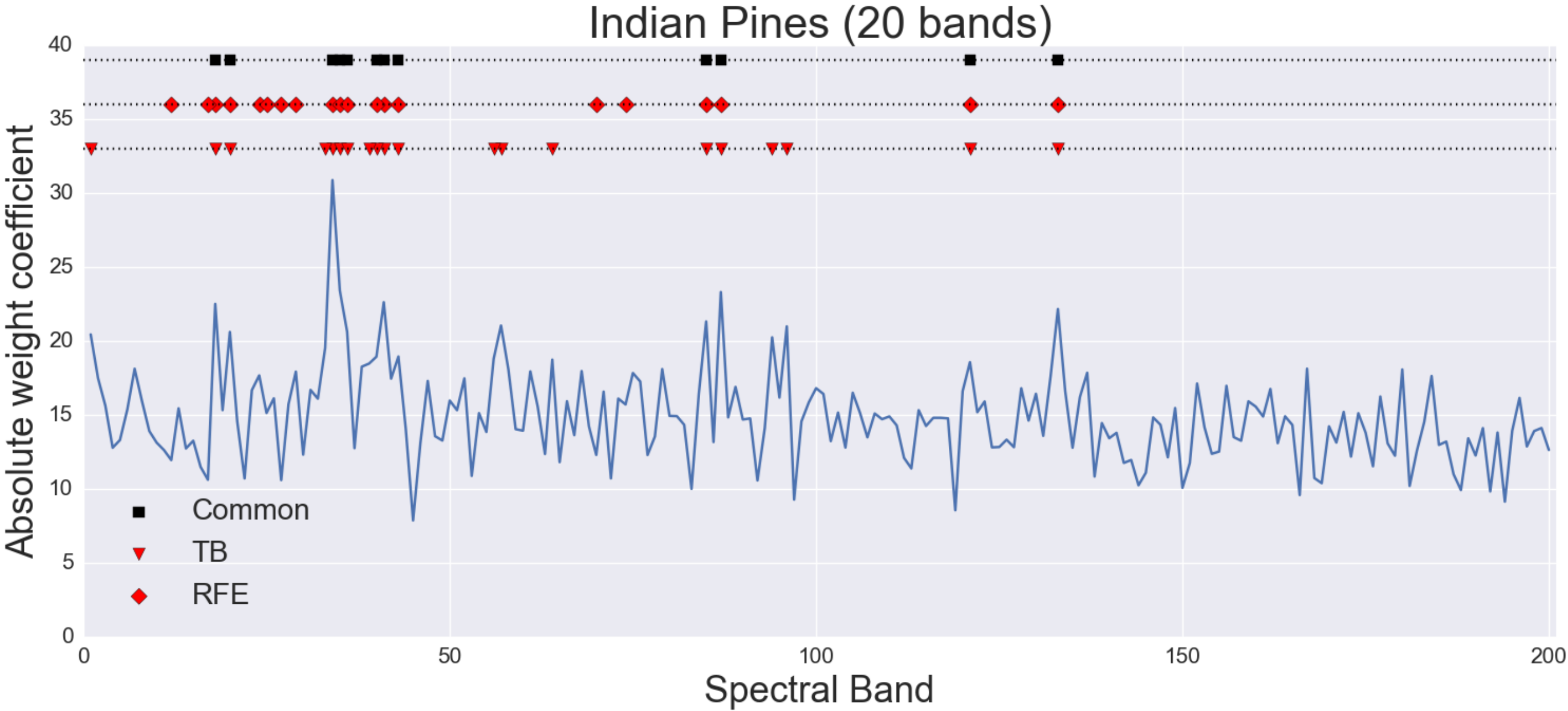}}
  \end{minipage}
  \begin{minipage}{0.5\linewidth}
    \centering
    \centerline{\includegraphics[width=1.0\linewidth]{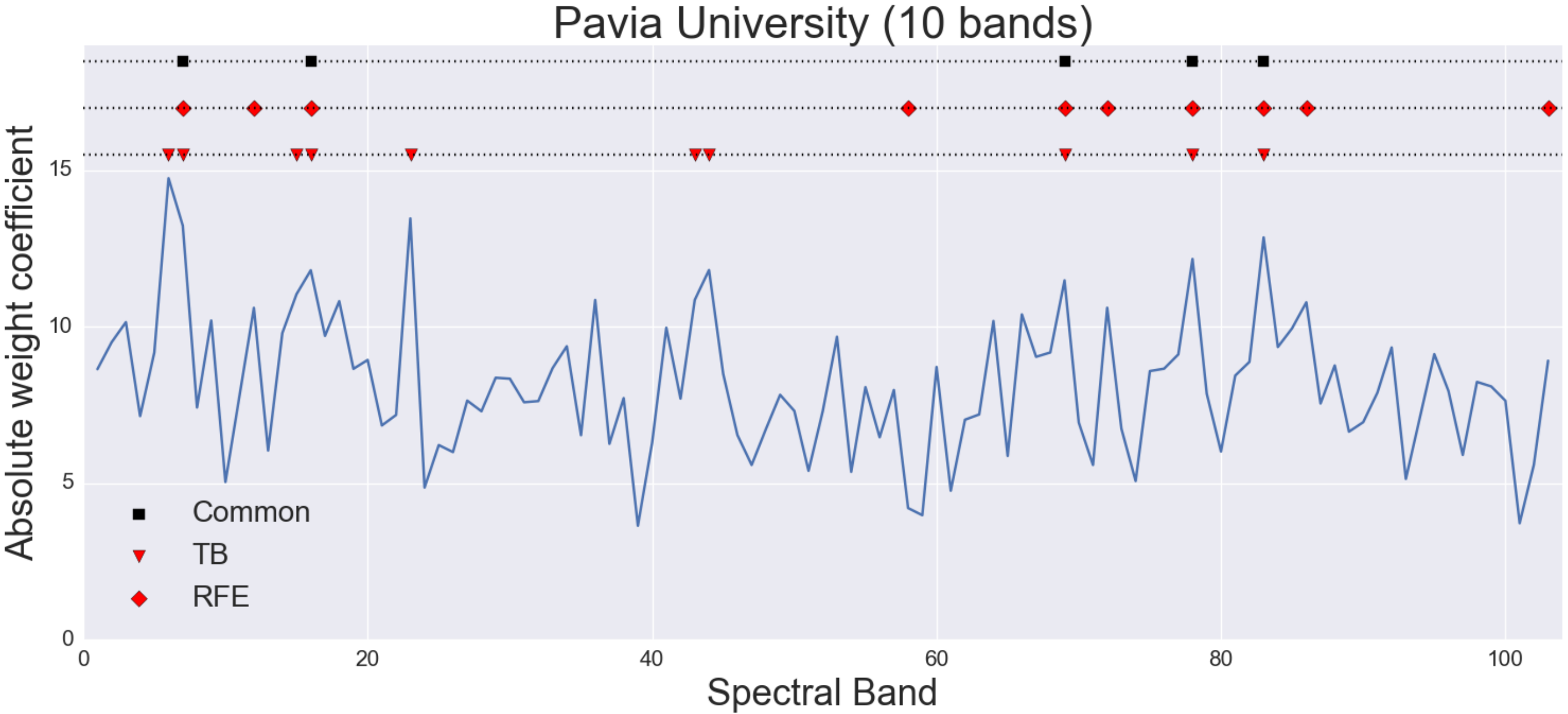}}
  \end{minipage} 
  \begin{minipage}{0.5\linewidth}
    \centering
    \centerline{\includegraphics[width=1.0\linewidth]{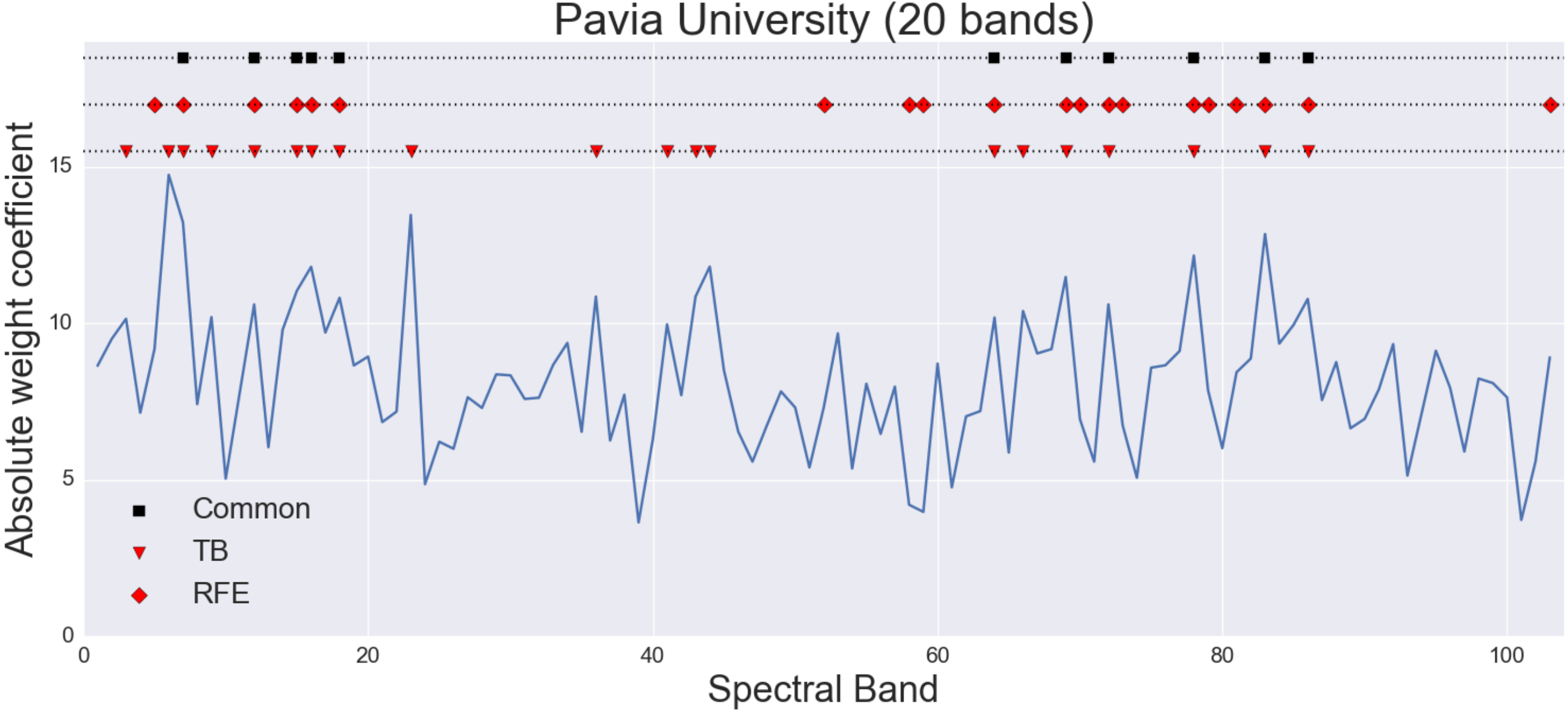}}
  \end{minipage}
  \caption{Selected spectral bands using the tensor-based (TB) logistic regression model and the RFE technique. Figures on the left depict the selection of the ten (10) most important bands, while figures on the right depict the selection of the twelve (20) most important bands. Common spectral bands of both techniques are also depicted, denoting as Common in the figure.}
  \label{fig:IP_important}
\end{figure*}

PCA-CNN performs slightly better than TB-CNN. However, our proposed method presents an advantage over PCA. Although, PCA maps the raw data to a lower dimensional feature space, the resulted features are not interpretable. Using the tensor-based dimensionality reduction, features at the lower dimensional space correspond to the most significant spectral bands providing a physical meaning on how spectral information affects the classification performance. 

\begin{figure}[t]
  \begin{minipage}{0.49\linewidth}
    \centering
    \centerline{\includegraphics[width=1.\linewidth]{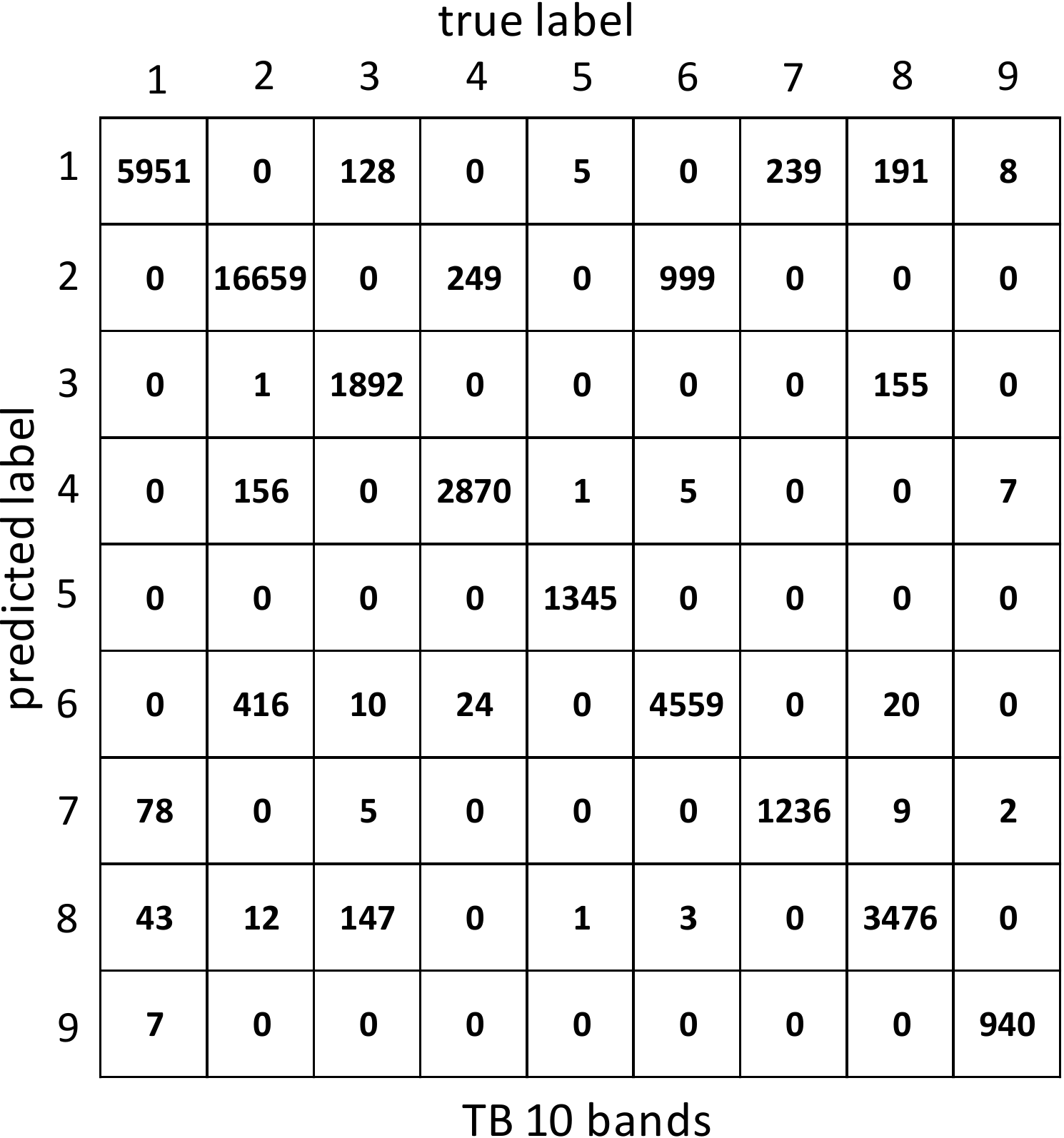}}
  \end{minipage} 
  \begin{minipage}{0.49\linewidth}
    \centering
    \centerline{\includegraphics[width=1.\linewidth]{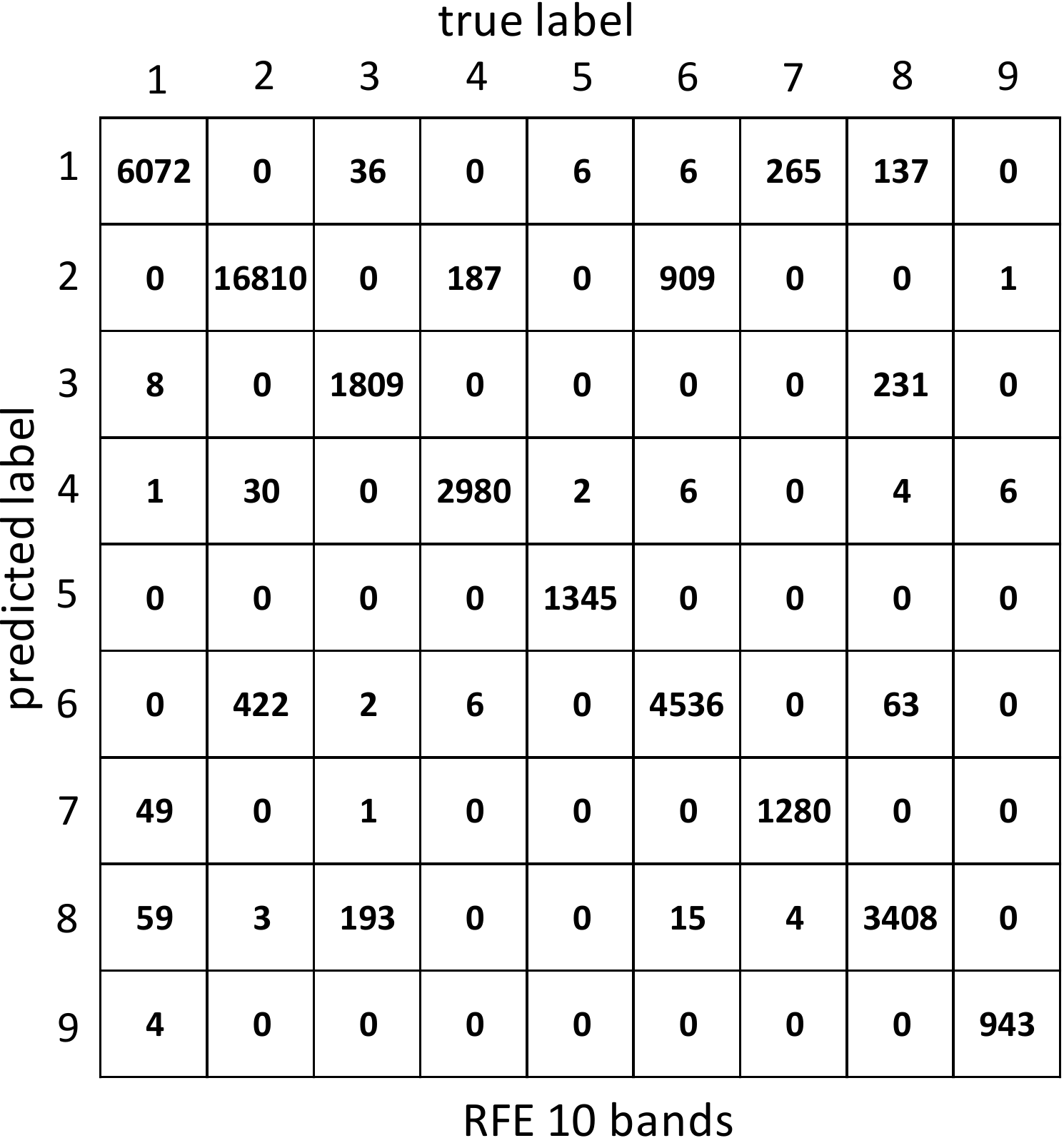}}
  \end{minipage} 
  
  \begin{minipage}{0.49\linewidth}
    \centering
    \centerline{\includegraphics[width=1.\linewidth]{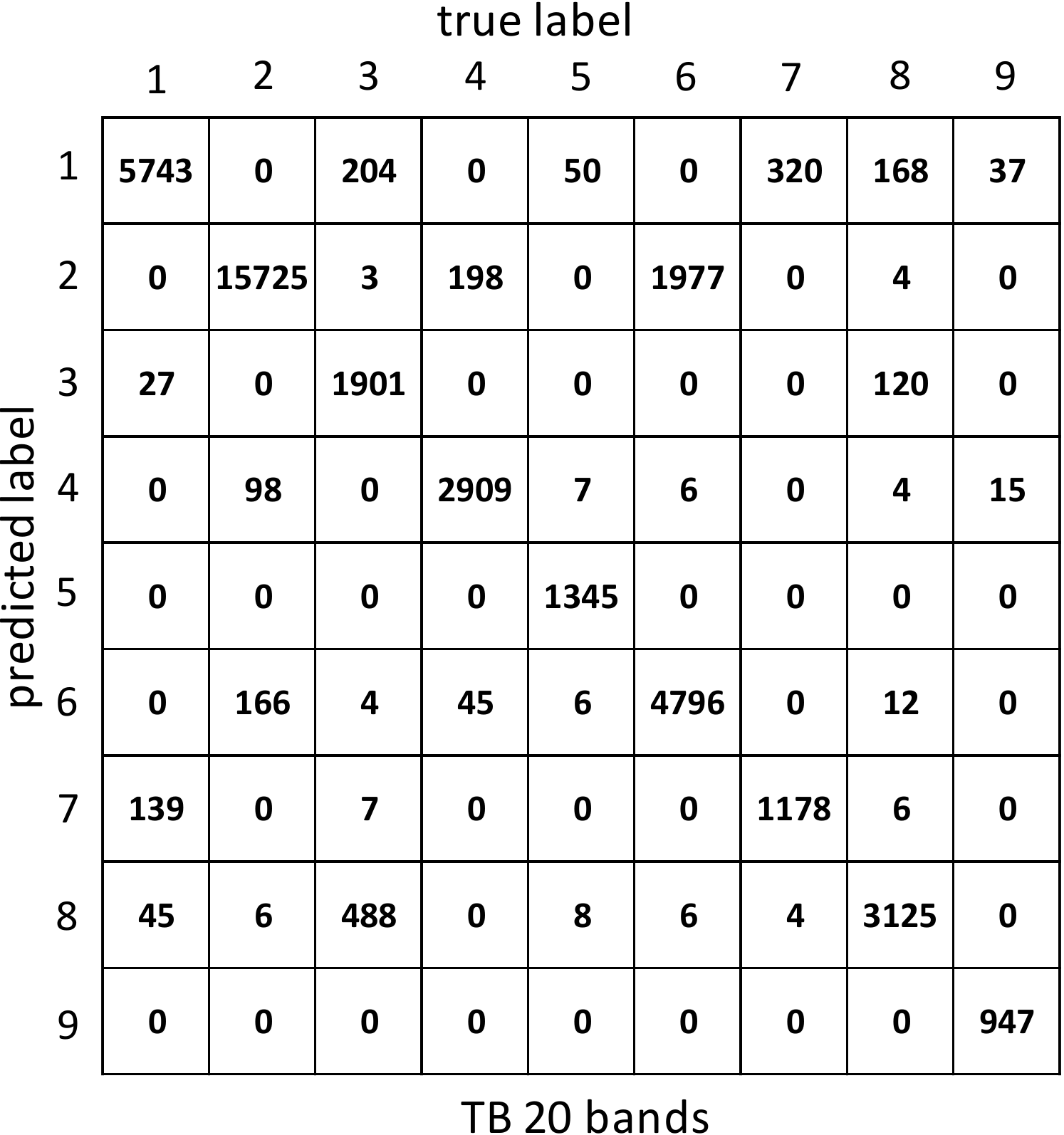}}
  \end{minipage} 
  \begin{minipage}{0.49\linewidth}
    \centering
    \centerline{\includegraphics[width=1.\linewidth]{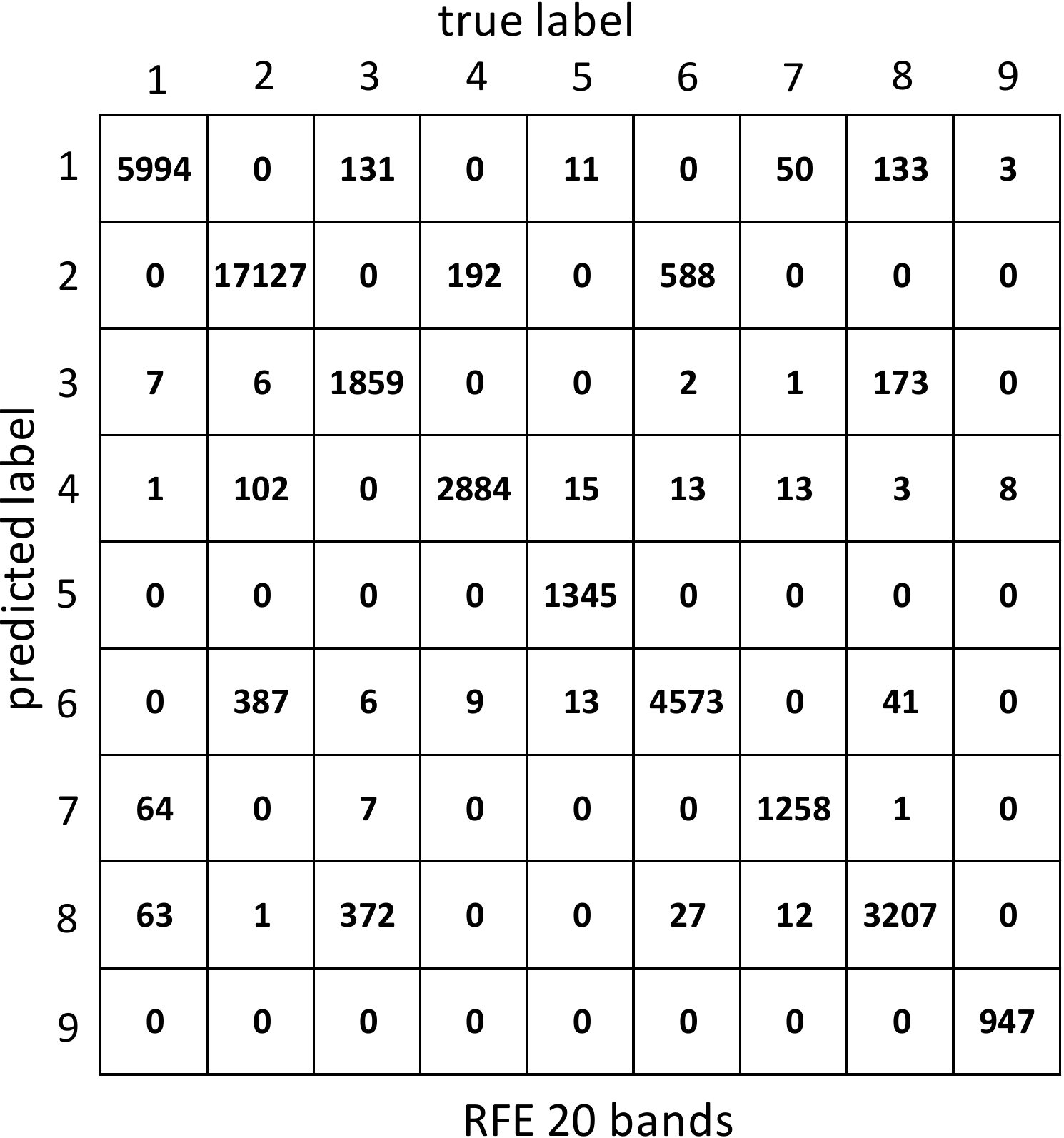}}
  \end{minipage} 
  
  \caption{Classification accuracy per class on Pavia University dataset when different feature selection techniques are used.}
  \label{fig:confusion_important}
\end{figure}

Furthermore, we compare the features extracted by the proposed tensor-based logistic regression model (high order linear modelling - Section \ref {sec:high_order_linear}) with the features extracted using the Recursive Feature Elimination (RFE) technique \cite{guyon2002gene}. RFE is an iterative procedure. At each iteration, it trains a model using all available spectral bands and then it discards the less informative ones until a predefined number of features is reached. Fig. \ref{fig:IP_important} presents the 10 and 20 most important spectral bands selected from both techniques (i.e., the tensor-based, called TB in the figure and the RFE). In this figure, the common bands detected by both techniques are also identified for clarification purposes. The results have been conducted for the India Pines and Pavia University dataset. To verify the quality of the selected spectral bands as important, we measure the classification accuracy obtained if we keep only these bands. Specifically, the extracted bands of both techniques are used to train the CNN model presented in \cite{Makantasis-etal:15}. Please note that the training set used for both feature selection techniques (i.e., TB and RFE) is the same and it was built by selecting 100 random samples per class. The results are presented in Table \ref{table:dimesionality_reduction} for Pavia dataset. In addition,  Fig. \ref{fig:confusion_important} shows the confusion matrix for the Pavia dataset when 10 or 20 salient bands are selected for the TB and RFE approach. The overall classification accuracy for both dimensionality reduction techniques is almost the same, indicating that the proposed TB method is also appropriate for dimensionality reduction apart from classification capabilities - see Section \ref{sub:tensor_based_logistic_regression}.

\begin{table*}[t]
\centering
\caption{Overall classification accuracy results (\%) when tdla and tensor-based logistic regression are used to reduce the dimension of the data.}
\newcolumntype{L}[1]{>{\hsize=#1\hsize\raggedright\arraybackslash}X}%
\newcolumntype{C}[1]{>{\hsize=#1\hsize\centering\arraybackslash}X}%
\label{table:TDLA}

\begin{tabularx}{1.0\linewidth}{L{1.}C{0.7}C{0.7}C{1.2}C{1.}C{1.2}C{1.}C{1.2}C{1.}}
\hline \hline 

DR + Classifier & TDLA + FCFFNN & TDLA + SVM & TB + FCFFNN (10 bands) & TB + SVM (10 bands)  & TB + FCFFNN (15 bands) & TB + SVM (15 bands) & TB + FCFFNN (20 bands) & TB + SVM (20 bands)\\ \hline

OA (\%) & 71.22 & 89.54 & 64.53 $\pm$ 1.4 & 68.14 $\pm$ 0.6 & 73.28 $\pm$ 0.8 & 73.77 $\pm$ 0.7 & 74.62 $\pm$ 0.7 & 73.95 $\pm$ 0.6   \\ \hline
 \hline

\end{tabularx}
\end{table*}

The features extracted by the tensor-based logistic regression model are also compared against the features selected by Tensor Discriminative Locality Alignment (TDLA) \cite{zhang2013tensor}, a state-of-the-art method for tensor data dimensionality reduction (see Table \ref{table:TDLA}). At this point, we should mention that TDLA has been designed for dimensionality reduction, while our method is a tensor-based classification modeling framework. Feature selection is an interesting side effect of our proposed methodology.

Again the quality of the data dimensionality reduction is measured through classification accuracy. In this experiment we have used different classification models, instead of the CNN adopted previously, to indicate the robustness of the proposed method in selecting salient spectral bands under different classification frameworks. Particularly, the experiments have been conducted using a FCFFNN and a SVM classifier which receives as inputs the selected salient bands of the TB and TDLA approach. We have used different classification models (e.g., CNN, FCFFNN, SVM) to indicate the robustness of the proposed method in selecting salient spectral bands under different classification frameworks. 
Three different experiments, for 10, 15, 20 most important spectral bands selection, are conducted. The experiments have been executed using 10 different runs and the standard deviation across all runs is depicted to indicate the robustness of the classification accuracy. As is observed, the best overall classification accuracy is achieved by TDLA and an SVM classifier. However, our approach gives better results when non-linear models (such as FCFFNN) are adopted.

The conclusions of the aforementioned experiments are that, although the proposed method has been designed for classification purposes (see the respective experiments of Section \ref{classification_framework}), it also performs quite well for data dimensionality reduction, compared with state of the art methods that have been appropriately designed for this purpose.

\begin{table}[t]
\centering
\caption{Classification accuracy results (\%) when RFE and tensor-based logistic regression are used to reduce the dimension of the data.}
\newcolumntype{L}[1]{>{\hsize=#1\hsize\raggedright\arraybackslash}X}%
\newcolumntype{C}[1]{>{\hsize=#1\hsize\centering\arraybackslash}X}%
\label{table:2}

\begin{tabularx}{0.98\linewidth}{C{5.5}C{4.5}C{8.0}C{6.7}}
\hline \hline 
\textbf{Feature selection} & \textbf{Number of bands} & \textbf{Accuracy (\%) Pavia University} & \textbf{Accuracy (\%) Indian Pines} \\ \hline
Tensor-based  & 10 & 92.82 & 77.48  \\ \hline
RFE           & 10 & 92.61 & 83.42  \\ \hline
Tensor-based  & 20 & 93.02 & 77.46  \\ \hline 
RFE           & 20 & 93.25 & 83.56  
  \\ \hline \hline \\
\label{table:dimesionality_reduction}
\end{tabularx}
\end{table}
\subsection{High-Order Nonlinear Classification Model}
\label{classification_framework}
\begin{figure*}[t]
  \begin{minipage}{0.5\linewidth}
    \centering
    \centerline{\includegraphics[width=0.95\linewidth]{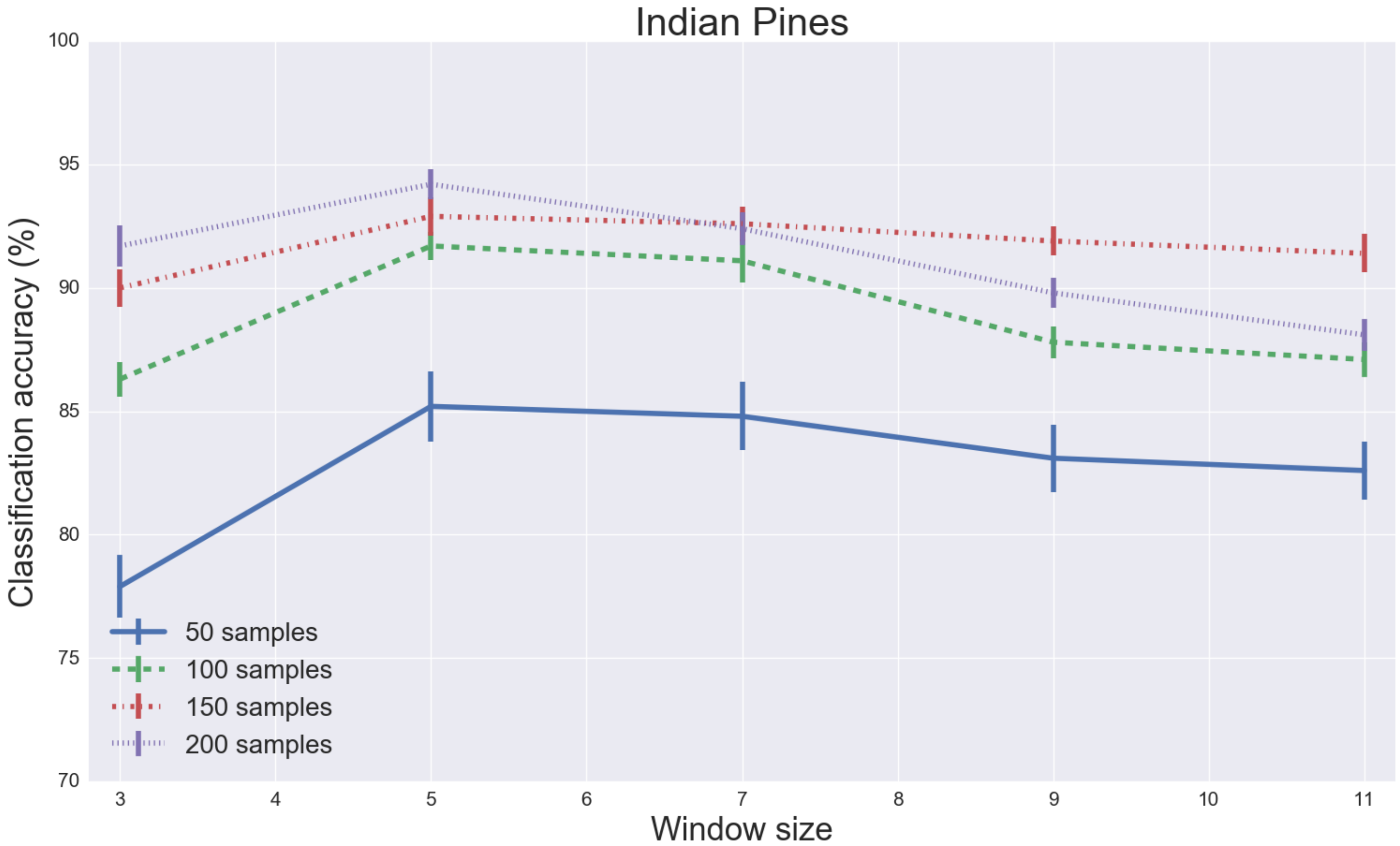}}
  \end{minipage} 
  \begin{minipage}{0.5\linewidth}
    \centering
    \centerline{\includegraphics[width=0.95\linewidth]{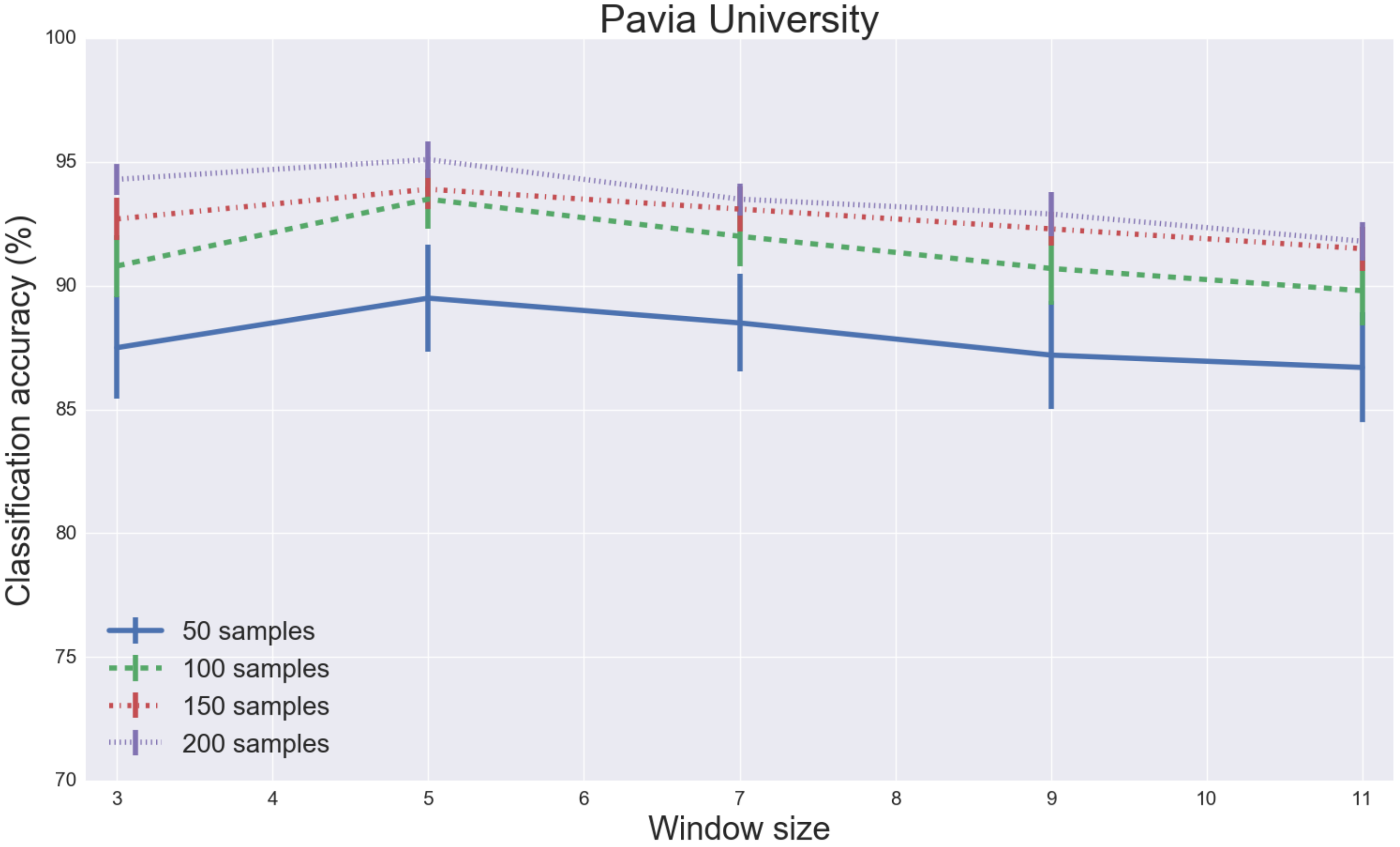}}
  \end{minipage}
  \caption{Overall classification accuracy on test set versus the size of spatial window for the high-order nonlinear classification model.}
  \label{fig:spatial_window}
\end{figure*}

\begin{figure*}[t]
  \begin{minipage}{0.5\linewidth}
    \centering
    \centerline{\includegraphics[width=0.95\linewidth]{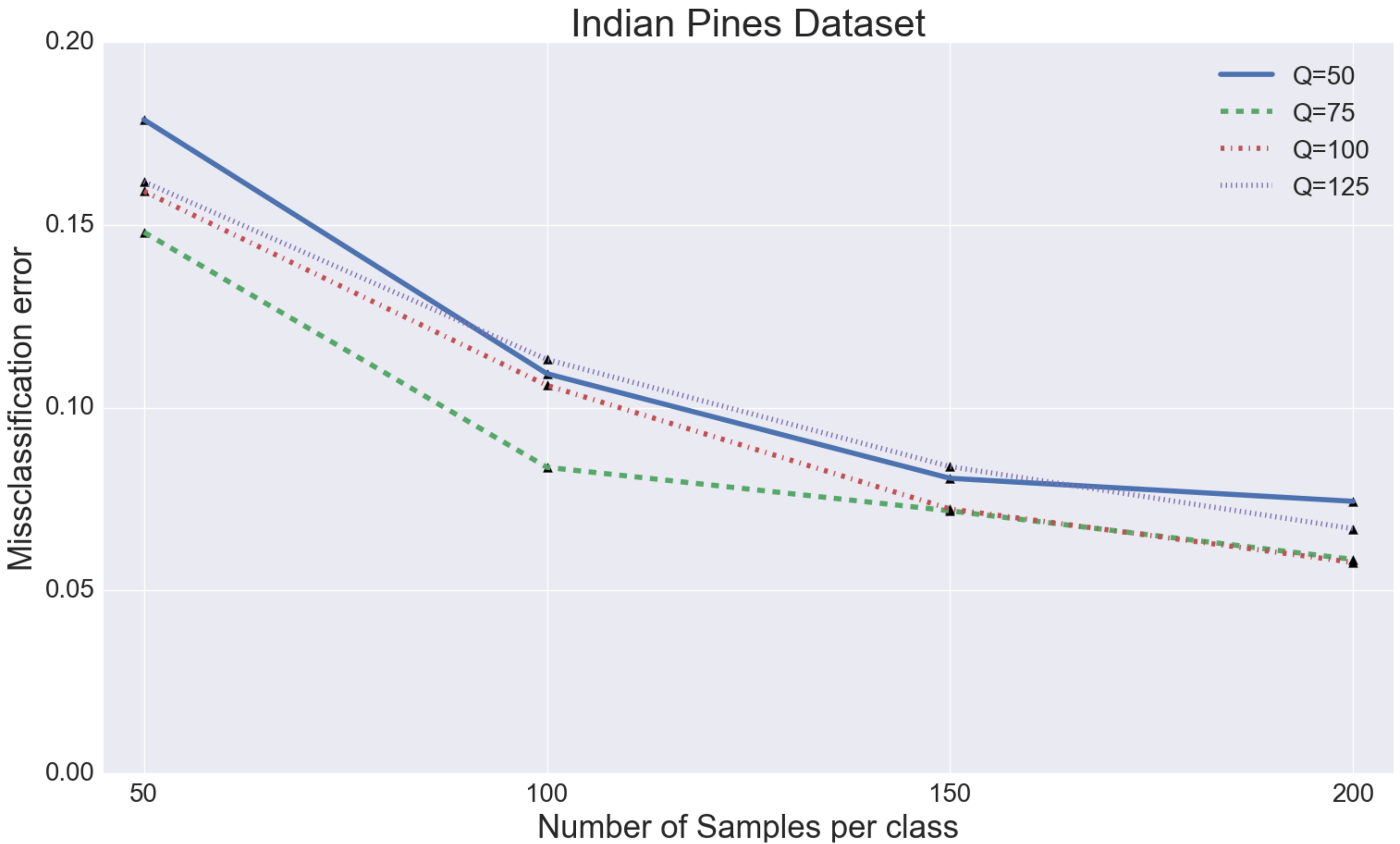}}
  \end{minipage} 
  \begin{minipage}{0.5\linewidth}
    \centering
    \centerline{\includegraphics[width=0.95\linewidth]{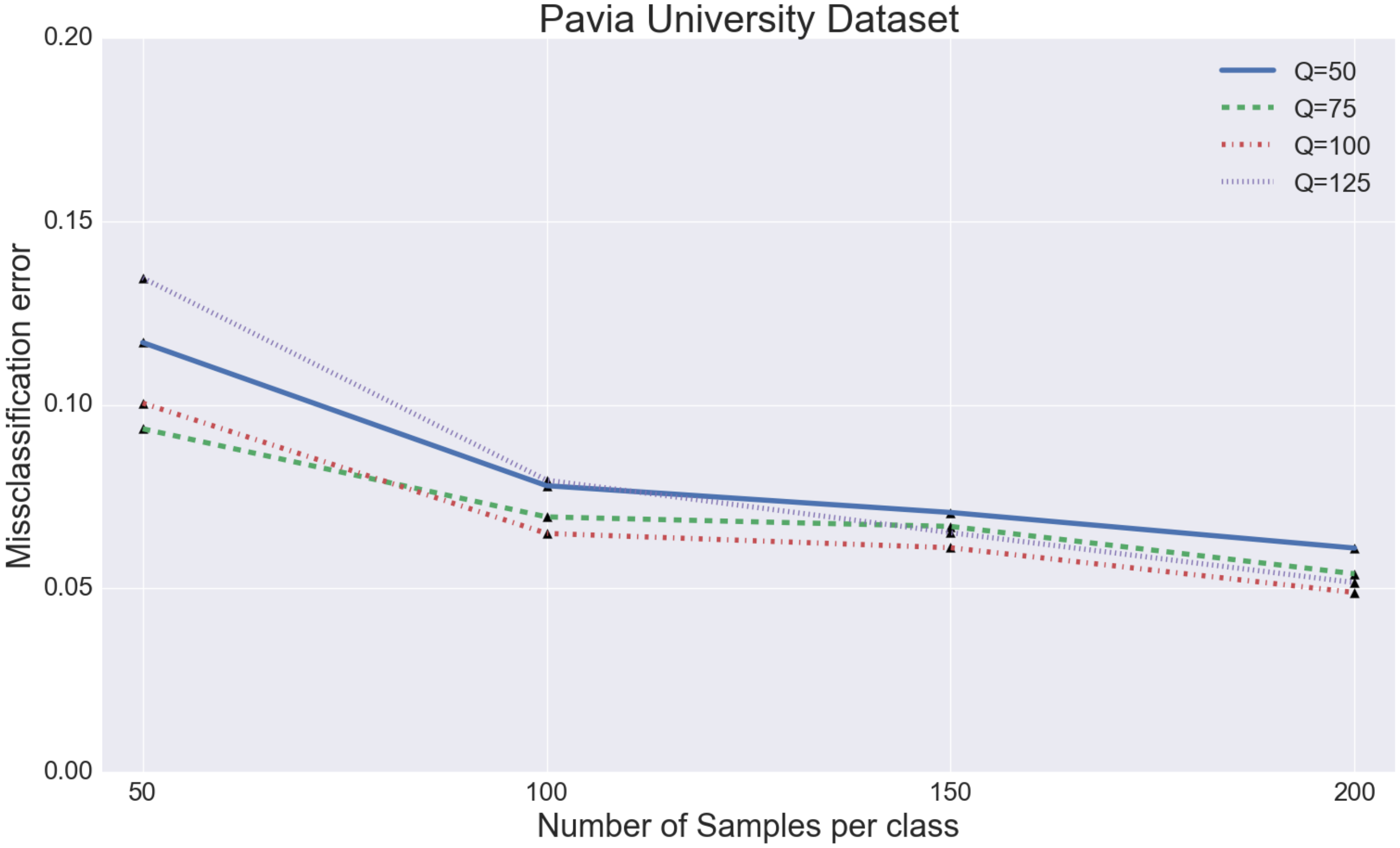}}
  \end{minipage}
  \caption{Misclassification error on test set versus the complexity, determined by $Q$, of the high-order nonlinear classification model.}
  \label{fig:2}
\end{figure*}

For evaluating the performance of the high-order nonlinear classification model, that is, the {\it rank}-1 FNN, we also use the Pavia University and the Indian Pines datasets. A similar procedure as in Section \ref{sub:tensor_based_logistic_regression} is followed to form the training sets. We conduct different experiments using training sets of 50, 100, 150 and 200 random samples from each class respectively. A similar training map selection was made in \cite{Li20151592} for the Indian Pines and Pavia University dataset. Please note that if for a specific class the number of samples are less than the number required for the experiment, then we select randomly 50\% of the total available class samples. Fig. \ref{fig:spatial_window} depicts the effect of different training samples on classification accuracy for different window size $s$ and assuming a constant number of neurons in the hidden layer. Particularly, we set $Q=100$ for the Pavia dataset and $Q=75$ for the Indian Pines. These values are derived from descriptions below (see also Fig. \ref{fig:2}). Additionally, Fig. \ref{fig:2} presents the effect of the number of training samples on misclassification error using different number of neurons $Q$ for both datasets.

Then, we evaluate the performance of proposed tensor-based high-order nonlinear classifier in relation to the size $s$ of the spatial window around a pixel. We conduct experiments with a window size to be equal to 3, 5, 7, 9, and 11 as is depicted in Figure \ref{fig:spatial_window}.
These results suggest that the best overall classification accuracy is achieved when $s=5$, i.e., the classification accuracy is mostly affected by the 24 closest neighbors of a pixel. When $s<5$ the spatial information is not adequate for achieving highly accurate classification results, while for $s>5$ the neighbourhood region is very probable to contain pixels that belong to different classes thus deteriorating the classification accuracy.

Moreover, we evaluate the performance of the high-order nonlinear classifier in relation to its complexity, i.e. the value of parameter $Q$, indicating the number of hidden neurons. Particularly, we set $Q$ to be equal to 50, 75, 100 and 125 respectively. By increasing the value of $Q$, the complexity and the capacity of the model are also increased. The results of this evaluation are presented in Fig.\ref{fig:2}.

Regarding the Indian Pines dataset, we observe that the model with $Q=75$ outperforms all other models. When the training set size is very small, i.e. 50 samples per class, the model with $Q=50$ does not have the capacity to capture the statistical relation between the input and the output and thus underfits the data. On the other hand, the models with $Q=100$ and $Q=125$, due to their high complexity, they slightly overfit the data. As training set increases, the misclassification error for all model decreases. This happens because increasing the training set size it also increases the amount of information that can be used during training to estimate the coefficients of the models.

As far as, the Pavia University dataset is concerned, we observe that the model with $Q=100$ outperforms all other models, when the size of dataset is larger than 50 samples per class. When the training dataset size is 50 samples per class the model with $Q=75$ outperforms all other models. The model with $Q=125$ overfits the data, while the model with $Q=50$ underfits them. When the size of the training dataset increases, the classification accuracy of all models also  increases. 

In the following, we compare the performance of the proposed \textit{rank}-1 FNN against FCFFNN, Radial Basis Function SVM (RBF-SVM), and two deep learning approaches that have been proposed for classifying hyperspectral data; the first one is based on Stacked-Autoencoders (SAE) \cite{chen2014deep}, while the second one on the exploitation of Convolutional Neural Networks (CNN) \cite{Makantasis-etal:15}. The FCFFNN receives as input the vectorized version of the Rank-1 FNN inputs. The same applies for the input of RBF-SVM. Moreover, FCFFNN has one hidden layer that has the same number of hidden neuron as the Rank-1 FNN. SAE and CNN were developed as proposed in the original papers. The number of hidden neurons of FCFFNN is 75 for the Indian Pines dataset and 100  the Pavia University dataset (as derived from Fig.\ref{fig:2}). The architecture of the network that exploits Stacked-Autoencoders consists of three hidden layers, while each hidden layer contains 10\% less neurons than its input. We choose to gradually reduce the number of hidden neurons from one hidden layer to the next, in order not to permit for the network to learn the identity function during pre-training. Regarding CNN, we utilize exactly the same architecture as the one presented in \cite{Makantasis-etal:15}. The performance of all these models is evaluated on varying size training sets; training sets that contain 50, 100, 150 and 200 samples from each one of the available classes.

\begin{table}[t]
\centering
\caption{Overall classification accuracy results (\%) of high-order nonlinear classification model for both datasets.}
\newcolumntype{L}[1]{>{\hsize=#1\hsize\raggedright\arraybackslash}X}%
\newcolumntype{C}[1]{>{\hsize=#1\hsize\centering\arraybackslash}X}%
\label{table:2}

\begin{tabularx}{0.98\linewidth}{L{10.0}C{3.5}C{3.5}C{3.5}C{3.5}}
\hline \hline \\
\multicolumn{5}{c}{{\bf Pavia University}} \\ \hline 
Samples per class & 50 & 100 & 150 & 200 \\ \hline

{\it rank}-1 FNN (Q=100) & \textbf{89.95 $\pm$ 0.7} & \textbf{93.50 $\pm$ 0.4} & 93.89 $\pm$ 0.3 & 95.11 $\pm$ 0.3   \\ \hline
FCFFNN        		  & 67.79 $\pm$ 3.6 & 76.53 $\pm$ 2.7 & 78.48 $\pm$ 2.2 & 82.59 $\pm$ 2.1   \\ \hline
RBF-SVM       		  & 86.98 $\pm$ 1.6 & 88.99 $\pm$ 1.3 & 89.86 $\pm$ 1.2 & 91.82 $\pm$ 1.2   \\ \hline 
SAE           		  & 86.54 $\pm$ 2.1 & 91.90 $\pm$ 1.8 & 92.38 $\pm$ 1.4 & 93.29 $\pm$ 1.1   \\ \hline 
CNN           		  & 88.89 $\pm$ 0.6 & 92.74 $\pm$ 0.3 & \textbf{94.68 $\pm$ 0.2} & \textbf{95.89 $\pm$ 0.2}   \\ \hline \hline \\

\multicolumn{5}{c}{{\bf Indian Pines}} \\ \hline 
Samples per class & 50 & 100 & 150 & 200 \\ \hline

{\it rank}-1 FNN (Q=75)  & \textbf{85.20 $\pm$ 1.2} & \textbf{91.63 $\pm$ 0.8} & \textbf{92.82 $\pm$ 0.5} & 94.15 $\pm$ 0.4   \\ \hline
FCFFNN        		 & 73.88 $\pm$ 4.2 & 81.10 $\pm$ 3.9 & 84.14 $\pm$ 2.7 & 85.86 $\pm$ 2.5   \\ \hline
RBF-SVM       		 & 73.18 $\pm$ 2.3 & 77.86 $\pm$ 2.0 & 82.11 $\pm$ 1.5 & 84.99 $\pm$ 1.5   \\ \hline 
SAE           		 & 65.51 $\pm$ 4.4 & 70.66 $\pm$ 3.6 & 74.03 $\pm$ 3.6 & 76.49 $\pm$ 3.2   \\ \hline 
CNN           		 & 82.43 $\pm$ 0.9 & 85.48 $\pm$ 0.7 & 92.28 $\pm$ 0.2 & \textbf{94.81 $\pm$ 0.2}   \\ \hline \hline

\end{tabularx}
\end{table}

\begin{table}[t]
\centering
\caption{Overall classification accuracy results (\%) of STM and Rank-1 FNN for both datasets.}
\newcolumntype{L}[1]{>{\hsize=#1\hsize\raggedright\arraybackslash}X}%
\newcolumntype{C}[1]{>{\hsize=#1\hsize\centering\arraybackslash}X}%
\label{table:STM}

\begin{tabularx}{0.98\linewidth}{L{7}C{6}C{6}C{6}}
\hline \hline \\
\multicolumn{4}{c}{{\bf Indian Pines}} \\ \hline 

Method & STM & STM-MPCA & Rank-1 FNN \\ \hline

OA  & 62.6 $\pm$ 2.6 & 80.6 $\pm$ 1.9 & 73.9 $\pm$ 1.4    \\ \hline \hline \hline \\

\multicolumn{4}{c}{{\bf Pavia University}}  \\ \hline 

Method & STM & STM-MPCA & Rank-1 FNN \\ \hline

OA  & 79.5 $\pm$ 1.4 & 89.4 $\pm$ 0.5 & 88.2 $\pm$ 0.8  \\ \hline
 \hline

\end{tabularx}
\end{table}

\begin{figure*}[t]
  \begin{minipage}{1.0\linewidth}
    \centering
    \centerline{\includegraphics[width=0.95\linewidth]{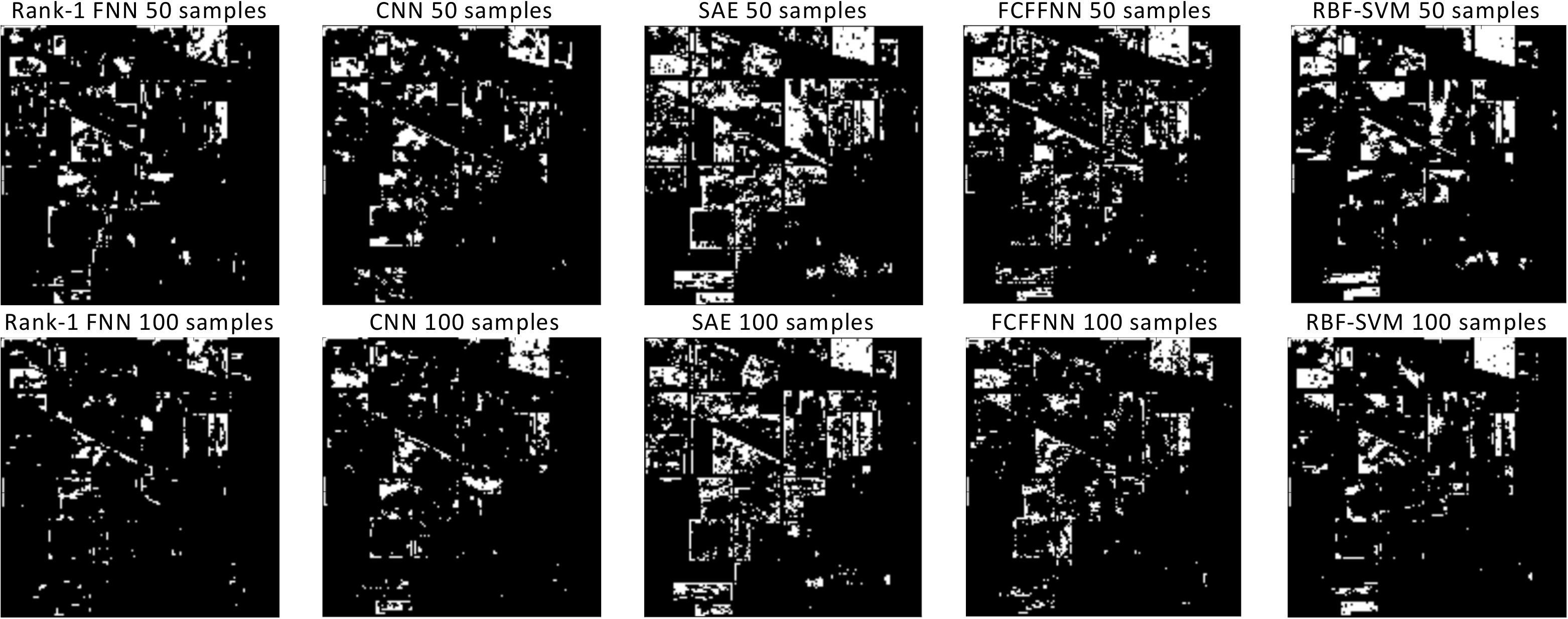}}
  \end{minipage}
  \caption{Visualization of classification accuracy for all tested approaches on Indian Pines dataset. White pixels were misclassified.} 
  \label{fig:3}
\end{figure*}

\begin{figure*}[t]
  \begin{minipage}{0.485\linewidth}
    \centering
    \centerline{\includegraphics[width=0.98\linewidth]{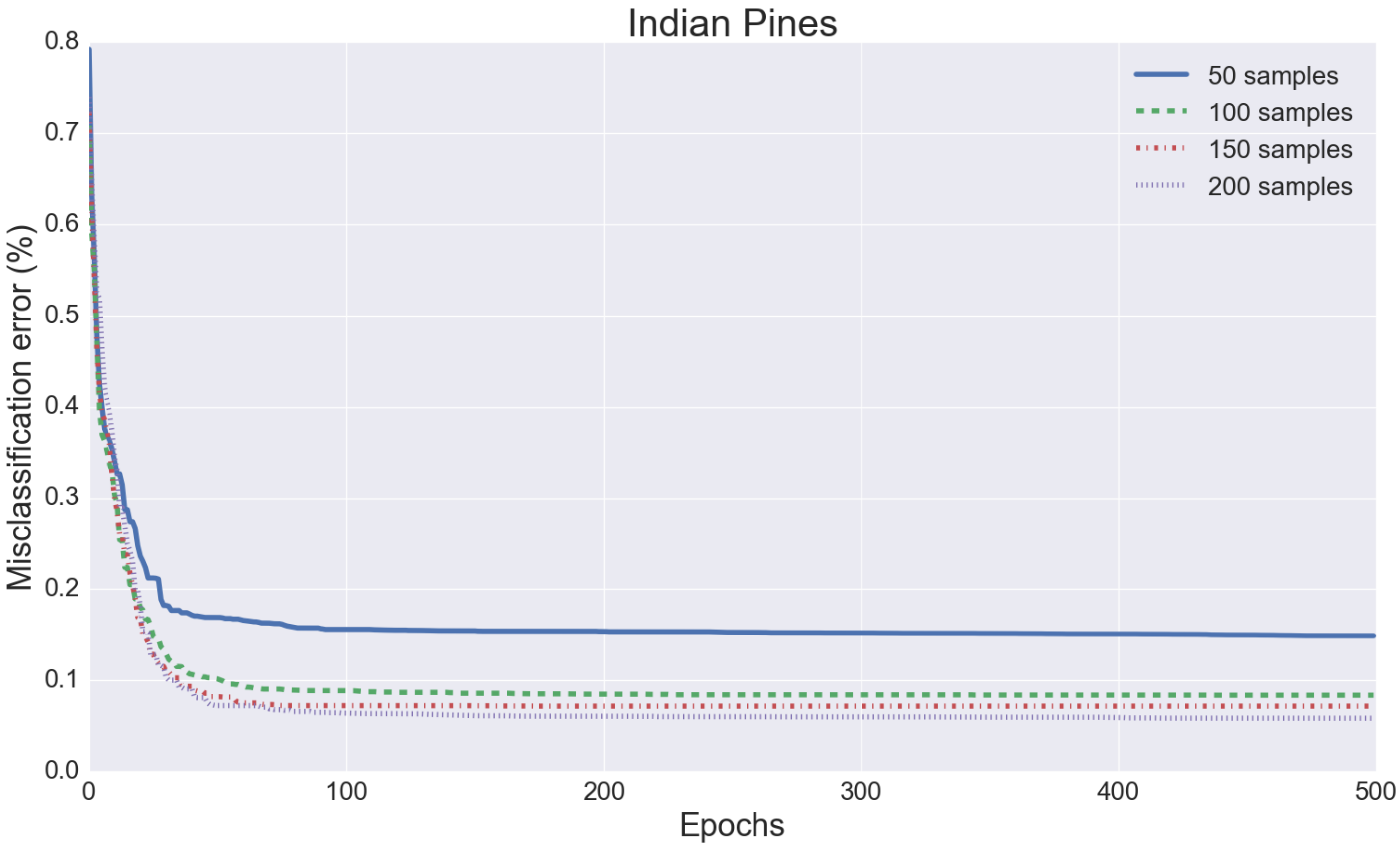}}
  \end{minipage} 
  \begin{minipage}{0.485\linewidth}
    \centering
    \centerline{\includegraphics[width=0.98\linewidth]{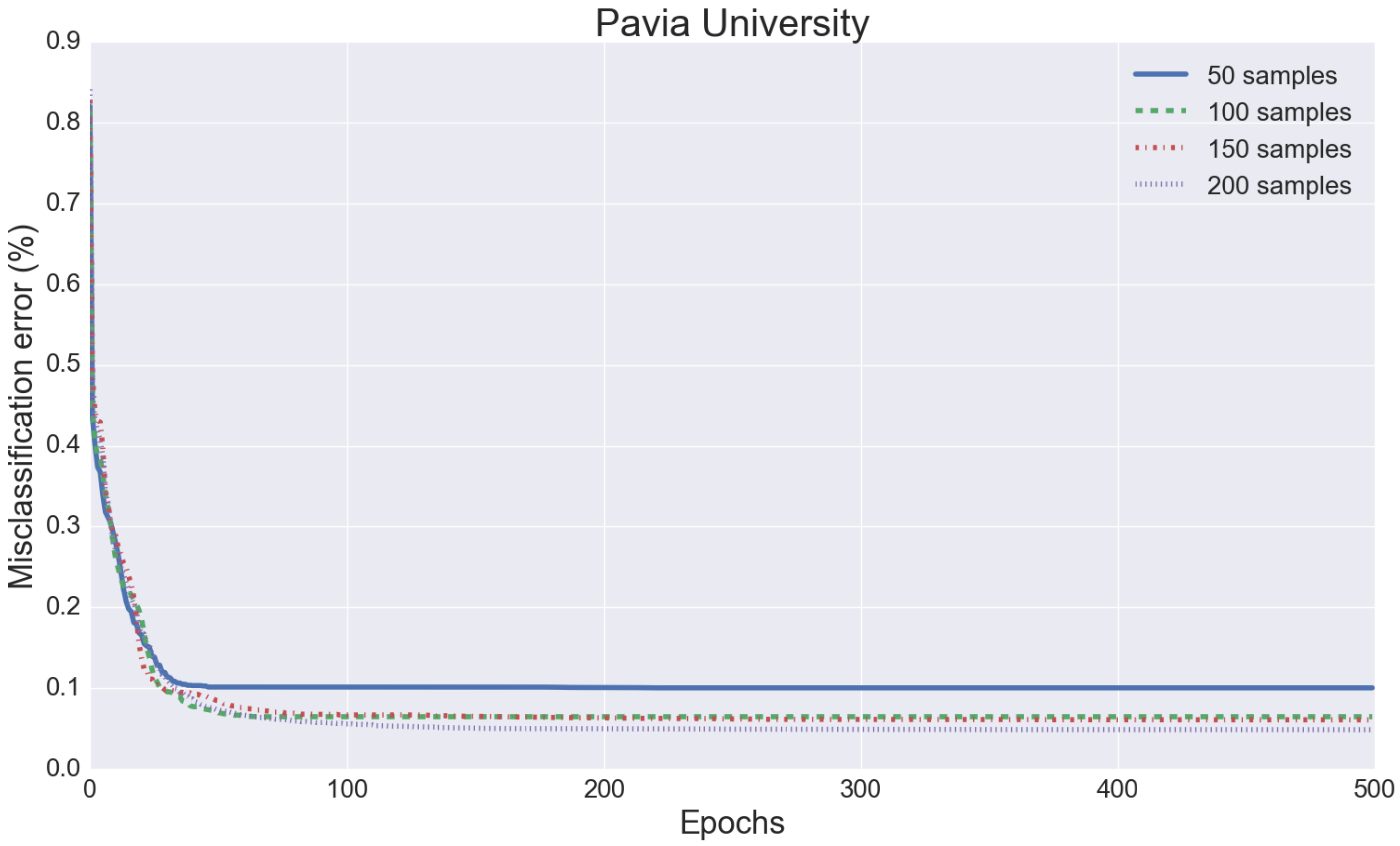}}
  \end{minipage}
  \caption{Convergence curves for the Rank-1 FNN for both datasets and for different sizes of training set.}
  \label{fig:convergence}
\end{figure*}

\begin{figure*}[t]
  \begin{minipage}{1.0\linewidth}
    \centering
    \centerline{\includegraphics[width=0.95\linewidth]{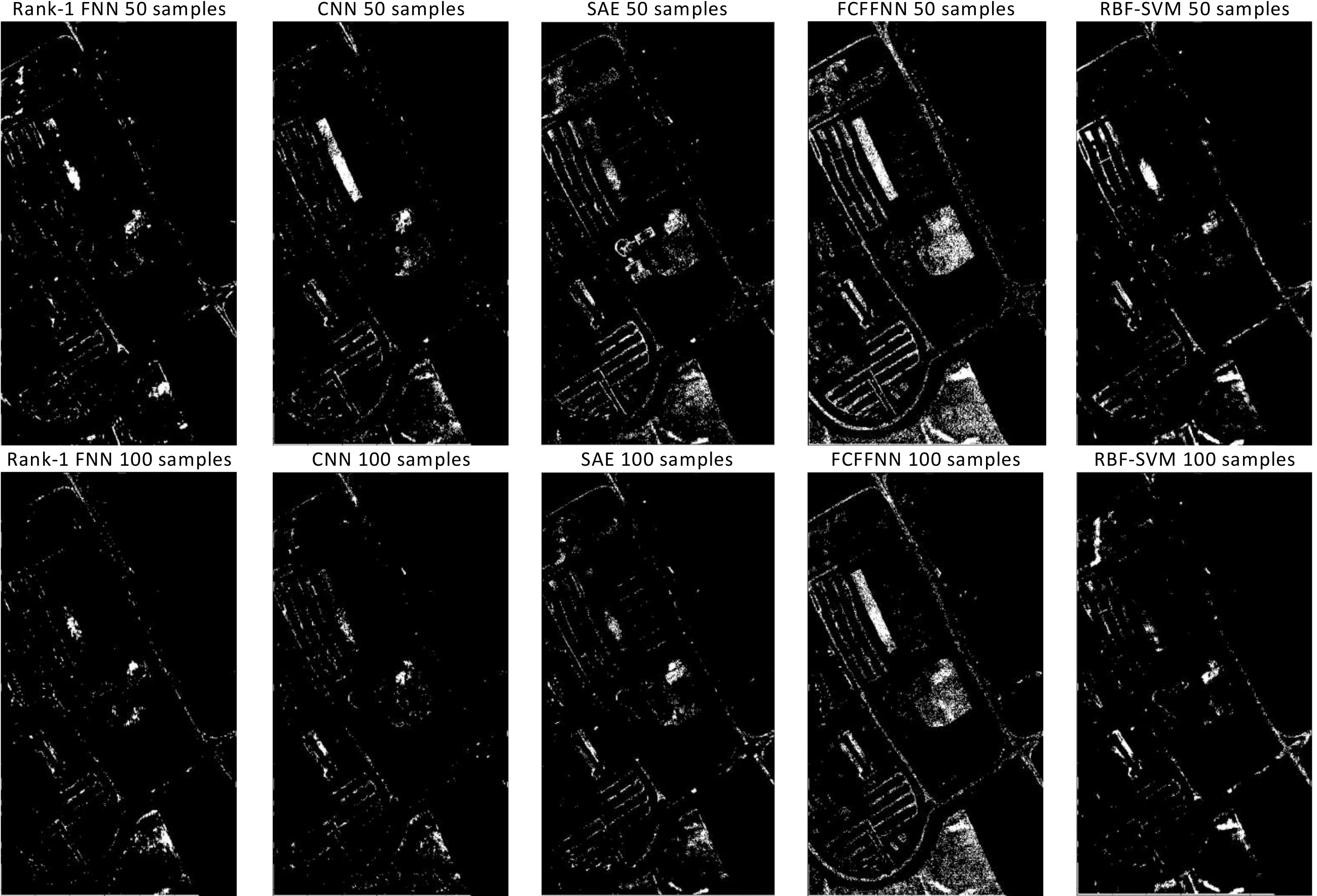}}
  \end{minipage}
  \caption{Visualization of classification accuracy for all tested approaches on Pavia University dataset. White pixels were misclassified.}
  \label{fig:4}
\end{figure*}

Table \ref{table:2} presents the outcome of this comparison. In this table, we also depict the standard deviation across 10 different experiment runs to beter reveal the classification accuracy. When the training set size is small, our approach outperforms all other models. This stems from the fact that the proposed high-order nonlinear model exploits tensor algebra operations to reduce the number of coefficients that need to be estimated during training, while at the same time it is able to retain the spatial information of the input. Although, the FCFFNN utilizes the same number of hidden neurons as our proposed model, it seems to overfit training sets when a small size of data is used, due to the fact that it employs a larger number of coefficients. RBF-SVM performs better than the FCFFNN on the Pavia University dataset, but slightly worse on the Indian Pines dataset. The deep learning architecture based on the exploitation of SAE, is actually a FCFFNN with three hidden layers, which employs an unsupervised pre-training phase to initialize its coefficients. The fully connectivity property of this architecture implies very high complexity, which is responsible for the poor performance, due to overfitting  issues, on the Indian Pines dataset. Finally, the deep learning architecture based on CNN  performs better that FCFFNN, RBF-SVM and SAE mainly due to its sparse connectivity (low complexity) and the fact that it can inherently exploit the spatial information of the input. When the training set consists of 150 and 200 samples per class, for the Pavia University dataset, and 200 samples for the Indian Pines dataset, the deep learning architecture based on CNN seems to outperform even the proposed high-order nonlinear classification model. This happens because the CNN-based deep learning model has higher capacity than the proposed model, which implies that it is capable of capturing better the statistical relation between the input and the output, when the training set contains sufficient information. However, when the size of the training set is small, our proposed model, due to its lower complexity, outperforms the CNN-based deep learning model. Fig. \ref{fig:3} and Fig. \ref{fig:4} depict the classification accuracy for all tested models when the trainig set contains 50 and 100 samples per class (white labeled pixels correspond to misclassified samples), while Fig. \ref{fig:convergence} presents the convergence curves for the Rank-1 FNN. In these figures, the false images (misclassified pixels) are depicted. 


Finally, we compare our proposed method against the Support Tensor Machine (STM) proposed in \cite{guo2016support} using Indian Pines and Pavia University datasets. The work of \cite{guo2016support} evaluates the performance of STM using both raw data and data of a reduced dimension derived through a Multilinear Principle Component Analysis (MPCA) \cite{lu2008mpca}. The results of this comparison are presented in Table \ref{table:STM}. In this table, we also depict the standard deviation of the results. As is observed, the proposed Rank-1 FNN outperforms the STM model when raw hyperspectral data are used. Instead, MPCA-STM works slightly better than our approach. This is mainly due to the fact that better features are used as inputs.

\section{Conclusions}
\label{sec:conclusions}
In this work, we present a linear and a non-linear tensor-based scheme for hyperspectral data classication and analysis. The advantages of the presented model is that i) it reduces the number of parameters required for the classification and thus it reduces the respective number of training samples, ii) it provides a physical interpretation regarding the model coefficients on the classification output and iii) it retains the spatial and spectral coherency of the input samples. By utilizing tensor algebra operations, we introduce learning algorithms to train the tensor-based classifiers (linear and non-linear). The linear classifier is based on a modification of a logistic regression model, while the non-linear one on a modification of a feedforward neural network. Both the proposed models assume a {\it rank}-1 canonical decomposition of the weight parameters. For this reason, we properly modified existing learning schemes to train these models so as to keep the {\it rank}-1 canonical decomposition property. We call the new proposed non-linear classifier as {\it rank}-1 FNN. 

We have evaluated the performance of the presented model in terms of classification accuracy and for dimensionality reduction. As far as the dimensionality reduction is concerned, the tensor-based scheme allows selection of the most discriminative spectral bands. It produces an interpretable dimensionality reduction, which can be used with more complicated classifiers without deteriorating their performance. In terms of classification, the linear classifier outperforms conventional linear models such as logistic regression and SVM. 

However, the linear classifier is characterized by low capacity, since it produces classification rules that are linear in the input space. For this reason, in this paper, we have introduced nonlinear classification models the weights of which satisfy the {\it rank}-1 canocial decomposition property. We have also introduced suitable learning algorithms to train the new proposed non-linear model. The performance of the nonlinear model was compared against other nonlinear classifiers, including the state-of-the-art  deep  learning  classifiers.  The  main  results  are that  when  the  size  of  the  training  set  is  small,  our  newly proposed model  presents  superior  performance  against  the  compared methods.

As future work, one can evaluate the performance of the proposed tensor-based classifier on fused datasets that contain hyperspectral data and LiDAR data \cite{Pacifici}. In this case, we need to investigate on the extraction of new features that better model how LiDAR data are related with hyperspectral particularities \cite{Maltezos2017}. 

\appendices
\section{}
\subsection{Proof of relation (\ref{eq:observation1})}
\noindent \textit{Proof}: For simplicity, we omit the superscripts of $\bm w$. Let us denote as $\bm W$ the tensor
\begin{equation}
    \bm W = \bm w_1 \circ \cdots \circ \bm w_D \in \mathbb R^{p_1,...,p_D} \nonumber
\end{equation}
having the same dimensions with the tensor $\bm X$. Then,
\begin{equation}
    vec(\bm W) = \bm w_1 \odot \cdots \odot \bm w_D. \nonumber
\end{equation}
We also have that
\begin{equation}
\begin{split}
    \langle \bm w_1 \odot \cdots & \odot \bm w_D, \bm X\rangle = \\ & \sum_{i_1}^{p_1} \cdots \sum_{i_D}^{p_D} W_{i_1,...,i_D}X_{i_1,...,i_D} = \\ & \langle \bm W_{(d)}, \bm X_{(d)} \nonumber
\end{split}
\end{equation}
for any $d = 1,2,...,D$. Furthermore, from relation (\ref{eq:3}) of the main manuscript, we have 
\begin{equation}
\label{eq:A1}
\begin{split}
   & \langle \bm W_{(d)}, \bm X_{(d)}\rangle = \\ & \langle \bm w_d (\bm w_D \odot \cdots \odot \bm w_{d+1} \odot \bm w_{d-1} \odot \cdots \odot \bm w_{1})^T, \bm X_{(d)} \rangle. 
\end{split}
    \tag{A.1}
\end{equation}
Now by using (\ref{eq:A1}) and the property of inner and dot products, which states that for any three matrices $A$, $B$ and $C$, the following 
\begin{equation}
    \langle \bm A \bm B^T, C \rangle = \bm A \bm B^T \bm C^T = \langle \bm A, \bm C \bm B \rangle \nonumber
\end{equation}
holds, we conclude that 
\begin{equation}
\begin{split}
    &\langle \bm w_1 \odot \cdots \odot \bm w_D, \bm X \rangle = \\
	 &\langle \bm w_d, \bm X_{(d)}(\bm w_D\odot \cdots \odot \bm w_{d+1}\odot \bm w_{d-1}\odot \cdots \odot \bm w_1) \rangle \nonumber.
\end{split}
\end{equation}
The proof is completed. $\lhd$

\bibliographystyle{IEEEtran}
\bibliography{IEEEabrv,refs}

\begin{thebibliography}{10}
\providecommand{\url}[1]{#1}
\csname url@samestyle\endcsname
\providecommand{\newblock}{\relax}
\providecommand{\bibinfo}[2]{#2}
\providecommand{\BIBentrySTDinterwordspacing}{\spaceskip=0pt\relax}
\providecommand{\BIBentryALTinterwordstretchfactor}{4}
\providecommand{\BIBentryALTinterwordspacing}{\spaceskip=\fontdimen2\font plus
\BIBentryALTinterwordstretchfactor\fontdimen3\font minus
  \fontdimen4\font\relax}
\providecommand{\BIBforeignlanguage}[2]{{%
\expandafter\ifx\csname l@#1\endcsname\relax
\typeout{** WARNING: IEEEtran.bst: No hyphenation pattern has been}%
\typeout{** loaded for the language `#1'. Using the pattern for}%
\typeout{** the default language instead.}%
\else
\language=\csname l@#1\endcsname
\fi
#2}}
\providecommand{\BIBdecl}{\relax}
\BIBdecl

\bibitem{wycoff2013non}
E.~Wycoff, T.-H. Chan, K.~Jia, W.-K. Ma, and Y.~Ma, ``A non-negative sparse
  promoting algorithm for high resolution hyperspectral imaging,'' in
  \emph{Acoustics, Speech and Signal Processing (ICASSP), 2013 IEEE
  International Conference on}.\hskip 1em plus 0.5em minus 0.4em\relax IEEE,
  2013, pp. 1409--1413.

\bibitem{chang2013hyperspectral}
C.-I. Chang, \emph{Hyperspectral data processing: algorithm design and
  analysis}.\hskip 1em plus 0.5em minus 0.4em\relax John Wiley \& Sons, 2013.

\bibitem{camps2005kernel}
G.~Camps-Valls and L.~Bruzzone, ``Kernel-based methods for hyperspectral image
  classification,'' \emph{IEEE Transactions on Geoscience and Remote Sensing},
  vol.~43, no.~6, pp. 1351--1362, 2005.

\bibitem{camps2009kernel}
------, \emph{Kernel methods for remote sensing data analysis}.\hskip 1em plus
  0.5em minus 0.4em\relax John Wiley \& Sons, 2009.

\bibitem{benitez1997artificial}
J.~M. Ben{\'\i}tez, J.~L. Castro, and I.~Requena, ``Are artificial neural
  networks black boxes?'' \emph{IEEE Transactions on neural networks}, vol.~8,
  no.~5, pp. 1156--1164, 1997.

\bibitem{camps2014advances}
G.~Camps-Valls, D.~Tuia, L.~Bruzzone, and J.~Atli~Benediktsson, ``Advances in
  hyperspectral image classification: Earth monitoring with statistical
  learning methods,'' \emph{Signal Processing Magazine, IEEE}, vol.~31, no.~1,
  pp. 45--54, 2014.

\bibitem{lecun1998gradient}
Y.~LeCun, L.~Bottou, Y.~Bengio, and P.~Haffner, ``Gradient-based learning
  applied to document recognition,'' \emph{Proceedings of the IEEE}, vol.~86,
  no.~11, pp. 2278--2324, 1998.

\bibitem{hinton2006reducing}
G.~E. Hinton and R.~R. Salakhutdinov, ``Reducing the dimensionality of data
  with neural networks,'' \emph{science}, vol. 313, no. 5786, pp. 504--507,
  2006.

\bibitem{hinton2006fast}
G.~E. Hinton, S.~Osindero, and Y.-W. Teh, ``A fast learning algorithm for deep
  belief nets,'' \emph{Neural computation}, vol.~18, no.~7, pp. 1527--1554,
  2006.

\bibitem{bengio2007greedy}
Y.~Bengio, P.~Lamblin, D.~Popovici, H.~Larochelle \emph{et~al.}, ``Greedy
  layer-wise training of deep networks,'' \emph{Advances in neural information
  processing systems}, vol.~19, p. 153, 2007.

\bibitem{chen2014deep}
Y.~Chen, Z.~Lin, X.~Zhao, G.~Wang, and Y.~Gu, ``Deep learning-based
  classification of hyperspectral data,'' \emph{IEEE Journal of Selected topics
  in applied earth observations and remote sensing}, vol.~7, no.~6, pp.
  2094--2107, 2014.

\bibitem{Makantasis-etal:15}
K.~Makantasis, K.~Karantzalos, A.~Doulamis, and N.~Doulamis, ``{Deep Supervised
  Learning for Hyperspectral Data Classification through Convolutional Neural
  Networks},'' in \emph{IEEE International Geoscience and Remote Sensing
  Symposium (IGARSS 2015)}, July 2015.

\bibitem{vakalopoulou2015building}
M.~Vakalopoulou, K.~Karantzalos, N.~Komodakis, and N.~Paragios, ``Building
  detection in very high resolution multispectral data with deep learning
  features,'' in \emph{Geoscience and Remote Sensing Symposium (IGARSS), 2015
  IEEE International}.\hskip 1em plus 0.5em minus 0.4em\relax IEEE, 2015, pp.
  1873--1876.

\bibitem{papadomanolaki2016benchmarking}
M.~Papadomanolaki, M.~Vakalopoulou, S.~Zagoruyko, and K.~Karantzalos,
  ``Benchmarking deep learning frameworks for the classification of very high
  resolution satellite multispectral data,'' \emph{ISPRS Annals of
  Photogrammetry, Remote Sensing \& Spatial Information Sciences}, vol.~3,
  no.~7, 2016.

\bibitem{mnih2012learning}
V.~Mnih and G.~E. Hinton, ``Learning to label aerial images from noisy data,''
  in \emph{Proceedings of the 29th International Conference on Machine Learning
  (ICML-12)}, 2012, pp. 567--574.

\bibitem{makantasis2015deepmm}
K.~Makantasis, K.~Karantzalos, A.~Doulamis, and K.~Loupos, ``Deep
  learning-based man-made object detection from hyperspectral data,'' in
  \emph{International Symposium on Visual Computing}.\hskip 1em plus 0.5em
  minus 0.4em\relax Springer, 2015, pp. 717--727.

\bibitem{orr2003neural}
G.~B. Orr and K.-R. M{\"u}ller, \emph{Neural networks: tricks of the
  trade}.\hskip 1em plus 0.5em minus 0.4em\relax Springer, 2003.

\bibitem{Kandylakis}
Z.~Kandylakis, K.~Karantzalos, A.~Doulamis, and N.~Doulamis, ``Multiple object
  tracking with background estimation in hyperspectral video sequences,'' in
  \emph{IEEE Workshop on Hyperspectral Image and Signal Processing: Evolution
  in Remote Sensing (WHISPERS)}, 2015.

\bibitem{kussul2016deep}
N.~Kussul, A.~Shelestov, M.~Lavreniuk, I.~Butko, and S.~Skakun, ``Deep learning
  approach for large scale land cover mapping based on remote sensing data
  fusion,'' in \emph{Geoscience and Remote Sensing Symposium (IGARSS), 2016
  IEEE International}.\hskip 1em plus 0.5em minus 0.4em\relax IEEE, 2016, pp.
  198--201.

\bibitem{makantasis2015tunnel}
K.~Makantasis, E.~Protopapadakis, A.~Doulamis, N.~Doulamis, and C.~Loupos,
  ``Deep convolutional neural networks for efficient vision based tunnel
  inspection,'' in \emph{Intelligent Computer Communication and Processing
  (ICCP), 2015 IEEE International Conference on}.\hskip 1em plus 0.5em minus
  0.4em\relax IEEE, 2015, pp. 335--342.

\bibitem{zhou2013tensor}
H.~Zhou, L.~Li, and H.~Zhu, ``Tensor regression with applications in
  neuroimaging data analysis,'' \emph{Journal of the American Statistical
  Association}, vol. 108, no. 502, pp. 540--552, 2013.

\bibitem{tan2012logistic}
X.~Tan, Y.~Zhang, S.~Tang, J.~Shao, F.~Wu, and Y.~Zhuang, ``Logistic tensor
  regression for classification,'' in \emph{International Conference on
  Intelligent Science and Intelligent Data Engineering}.\hskip 1em plus 0.5em
  minus 0.4em\relax Springer, 2012, pp. 573--581.

\bibitem{de2004computation}
L.~De~Lathauwer, B.~De~Moor, and J.~Vandewalle, ``Computation of the canonical
  decomposition by means of a simultaneous generalized schur decomposition,''
  \emph{SIAM journal on Matrix Analysis and Applications}, vol.~26, no.~2, pp.
  295--327, 2004.

\bibitem{csaji2001approximation}
B.~C. Cs{\'a}ji, ``Approximation with artificial neural networks,''
  \emph{Faculty of Sciences, Etvs Lornd University, Hungary}, vol.~24, p.~48,
  2001.

\bibitem{kolda2009tensor}
T.~G. Kolda and B.~W. Bader, ``Tensor decompositions and applications,''
  \emph{SIAM review}, vol.~51, no.~3, pp. 455--500, 2009.

\bibitem{bishop2006pattern}
C.~M. Bishop, \emph{Pattern recognition and machine learning}.\hskip 1em plus
  0.5em minus 0.4em\relax springer, 2006.

\bibitem{hung2013matrix}
H.~Hung and C.-C. Wang, ``Matrix variate logistic regression model with
  application to eeg data,'' \emph{Biostatistics}, vol.~14, no.~1, pp.
  189--202, 2013.

\bibitem{harshman1970foundations}
R.~A. Harshman, ``Foundations of the parafac procedure: Models and conditions
  for an" explanatory" multi-modal factor analysis,'' 1970.

\bibitem{carroll1970analysis}
J.~D. Carroll and J.-J. Chang, ``Analysis of individual differences in
  multidimensional scaling via an n-way generalization of “eckart-young”
  decomposition,'' \emph{Psychometrika}, vol.~35, no.~3, pp. 283--319, 1970.

\bibitem{ndoulam}
A.~D. Doulamis, N.~D. Doulamis, and S.~D. Kollias, ``On-line retrainable neural
  networks: improving the performance of neural networks in image analysis
  problems,'' \emph{IEEE Transactions on Neural Networks}, vol.~11, no.~1, pp.
  137--155, 2000.

\bibitem{adoulam1}
A.~Doulamis, N.~Doulamis, K.~Ntalianis, and S.~Kollias, ``An efficient fully
  unsupervised video object segmentation scheme using an adaptive
  neural-network classifier architecture,'' \emph{IEEE Transactions on Neural
  Networks}, vol.~14, no.~3, pp. 616--630, 2003.

\bibitem{adoulam}
A.~D. Doulamis, N.~D. Doulamis, and S.~D. Kollias, ``An adaptable
  neural-network model for recursive nonlinear traffic prediction and modeling
  of mpeg video sources,'' \emph{IEEE Transactions on Neural Networks},
  vol.~14, no.~1, pp. 150--166, 2003.

\bibitem{guyon2002gene}
I.~Guyon, J.~Weston, S.~Barnhill, and V.~Vapnik, ``Gene selection for cancer
  classification using support vector machines,'' \emph{Machine learning},
  vol.~46, no. 1-3, pp. 389--422, 2002.

\bibitem{zhang2013tensor}
L.~Zhang, L.~Zhang, D.~Tao, and X.~Huang, ``Tensor discriminative locality
  alignment for hyperspectral image spectral--spatial feature extraction,''
  \emph{IEEE Transactions on Geoscience and Remote Sensing}, vol.~51, no.~1,
  pp. 242--256, 2013.

\bibitem{Li20151592}
J.~Li, X.~Huang, P.~Gamba, J.~Bioucas-Dias, L.~Zhang, J.~Benediktsson, and
  A.~Plaza, ``Multiple feature learning for hyperspectral image
  classification,'' \emph{IEEE Transactions on Geoscience and Remote Sensing},
  vol.~53, no.~3, pp. 1592--1606, 2015.

\bibitem{guo2016support}
X.~Guo, X.~Huang, L.~Zhang, L.~Zhang, A.~Plaza, and J.~A. Benediktsson,
  ``Support tensor machines for classification of hyperspectral remote sensing
  imagery,'' \emph{IEEE Transactions on Geoscience and Remote Sensing},
  vol.~54, no.~6, pp. 3248--3264, 2016.

\bibitem{lu2008mpca}
H.~Lu, K.~N. Plataniotis, and A.~N. Venetsanopoulos, ``Mpca: Multilinear
  principal component analysis of tensor objects,'' \emph{IEEE transactions on
  Neural Networks}, vol.~19, no.~1, pp. 18--39, 2008.

\bibitem{Pacifici}
F.~Pacifici, Q.~Du, and S.~Prasad, ``Report on the 2013 ieee grss data fusion
  contest: Fusion of hyperspectral and lidar data [technical committees],''
  \emph{IEEE Transactions on Geoscience and Remote Sensing}, vol.~1, no.~3, pp.
  136--38, 2013.

\bibitem{Maltezos2017}
E.~Maltezos, N.~Doulamis, A.~Doulamis, and C.~Ioannidis, ``Deep convolutional
  neural networks for building extraction from orthoimages and dense image
  matching point clouds,'' \emph{Journal of Applied Remote Sensing}, vol.~11,
  no.~4, 2017.

\end{thebibliography}

\vfill


\end{document}